\documentclass{llncs}
\usepackage{makeidx}  
\usepackage{amsmath} 
\usepackage{amssymb} 
\usepackage[linesnumbered,lined]{algorithm2e}
\usepackage{cite}
\usepackage{graphicx}
\usepackage{relsize}
\graphicspath{ {./ImgsIPMI_75/} }
\usepackage{array}
\usepackage{multirow}
\usepackage{rotating}
\usepackage{wasysym}
\usepackage{mathtools}
\usepackage{tabularx}
\usepackage{cleveref}
\usepackage{floatrow}
\usepackage[misc]{ifsym}
\begin{document}
\title{VTrails: Inferring Vessels with Geodesic Connectivity Trees}
\author{Stefano Moriconi\inst{1}\textsuperscript{(\Letter)} \and Maria A. Zuluaga\inst{1} \and H. Rolf J{\"a}ger\inst{2} \and Parashkev Nachev\inst{2} \and S\'ebastien Ourselin\inst{1,3} \and M. Jorge Cardoso\inst{1,3} }
\institute{Translational Imaging Group, CMIC, University College London, London, UK \email{stefano.moriconi.15@ucl.ac.uk}
\and
Institute of Neurology, University College London, London, UK
\and
Dementia Research Centre, University College London, London, UK}
\maketitle
\begin{abstract}
The analysis of vessel morphology and connectivity has an impact on a number of cardiovascular and neurovascular applications by providing patient-specific high-level quantitative features such as spatial location, direction and scale. In this paper we present an end-to-end approach to extract an acyclic vascular tree from angiographic data by solving a connectivity-enforcing anisotropic fast marching over a voxel-wise tensor field representing the orientation of the underlying vascular tree. The method is validated using synthetic and real vascular images. We compare VTrails against classical and state-of-the-art ridge detectors for tubular structures by assessing the connectedness of the vesselness map and inspecting the synthesized tensor field as proof of concept. VTrails performance is evaluated on images with different levels of degradation: we verify that the extracted vascular network is an acyclic graph (i.e. a tree), and we report the extraction accuracy, precision and recall.
\end{abstract}
\section{Introduction}
Vessel morphology and connectivity is of clinical relevance in cardiovascular and neurovascular applications. In clinical practice, the vascular network and its abnormalities are assessed by inspecting intensity projections, or image slices one at a time, or using multiple views of 3D rendering techniques. 
In a number of conditions, the connected vessel segmentation is required for intervention or treatment planning \cite{Zuluaga2015}. A schematic representation of the vascular network has an impact in interventional neuroradiology and in vascular surgery by providing patient-specific high-level quantitative features (spatial localization, direction and scale). In vascular image analysis these features are used for segmentation and labelling \cite{Lesage2009}, with the final aim of reconstructing a physical vascular model for hemodynamic simulations, or catheter motion planning, or identifying (un)safe occlusion points \cite{Bullitt1999}.
With this view, previous studies addressed the problem of extracting a connected vascular network in a disjoint manner. First, \cite{frangi1998multiscale, law2008three} proposed tubular enhancing methods in 3D with the aim of better contrasting vessels over a background: by using the eigendecomposition of either the Hessian matrix, or the image gradient projected on a unit sphere boundary, a scalar \textit{vesselness} measure is obtained, which represents a vascular saliency map. Secondly, given the vascular saliency map, local disconnected branches or fragmented centerlines, \cite{Bullitt1999, kwitt2013studying, schaap2007bayesian} proposed a set of methods to recover a connected network: `cores' identify and track furcating branches, whereas vascular graphs are recovered using minimum spanning tree algorithms on image-intensity features, or using graph kernels (subtree patterns) matched on a similarity metric. Alternatively, geometrical models embedding shape priors, or probabilistic models based on image-related features were employed to recover the connected vessel centerlines and prune artifacts from an initial set of segments. A different approach is proposed in \cite{Antiga2008}, where the connected centerlines are recovered \textit{a-posteriori} as medial axes of the 3D surface model which segments the lumen.\\
Given the varying complexity of the vascular network in healthy and diseased subjects and the lack of an extensive connected ground-truth for complex vascular networks of several anatomical compartments, the accurate and exhaustive extraction of the vessel connectivity remains however a challenging task.\\
Here we propose \textit{VTrails}, a novel method that addresses vascular connectivity under a unified mathematical framework. VTrails enhances the connectedness of furcating, fragmented and tortuous vessels through scalar and high-order vascular features, which are employed in a greedy connectivity paradigm to determine the final vascular network. In particular, the vascular image is filtered first with a Steerable Laplacian of Gaussian Swirls filterbank, synthesizing simultaneously a connected vesselness map and an associated tensor field.
Under the assumption that vessels join by minimal paths, VTrails then infers the unknown fully-connected vascular network as the minimal cost acyclic graph connecting automatically extracted seed nodes.
\section{Methods}
We introduce in section 2.1 a Steerable Laplacian of Gaussian Swirls \small (SLoGS) \normalsize filterbank used to reconstruct simultaneously the vesselness map and the associated tensor field. The \small SLoGS \normalsize filterbank is first defined, then a multiscale image filtering approach is described using a locally selective overlap-add method \cite{oppenheim2010discrete}. The connected vesselness map and the tensor field are integrated over scales.\\
In section 2.2, an anisotropic level-set combined with a connectivity paradigm extracts the fully-connected vascular tree using the synthesized connected vesselness map and tensor field.
\begin{figure}[tb!]
\begin{tabular}{@{}c@{}c|c@{}c@{}}
	\multicolumn{2}{c|}{\scriptsize{\textbf{Steerable Laplacian of Gaussian Swirls}}} & \multicolumn{2}{c}{\scriptsize{\textbf{Connected Vesselness and Tensor Field}}} \\
	\tiny{$\Gamma( \mathbf{x},\boldsymbol{\sigma} ,\mathbf{c})$} & \tiny{$\Gamma_{\text{tube}}$} & \tiny{Multi-scale Pyramid} & \\
	\multirow{3}{*}[1em]{
										\begin{tabular}{@{}c@{}c@{}}
											\raisebox{-.5\height}{\includegraphics[width=0.1\textwidth]{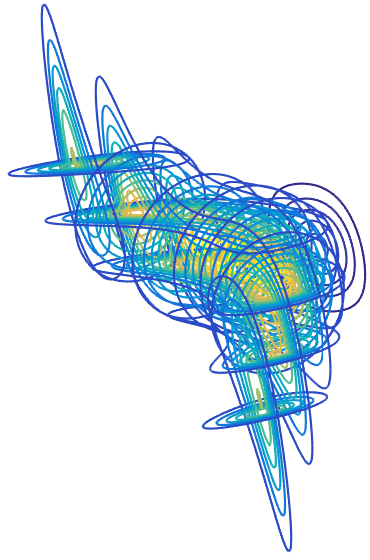}} & 
											\raisebox{-.5\height}{\includegraphics[width=0.1\textwidth]{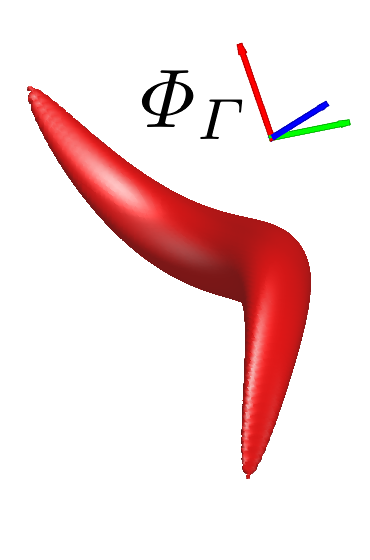}} \\
											\raisebox{-.5\height}{\includegraphics[width=0.09\textwidth]{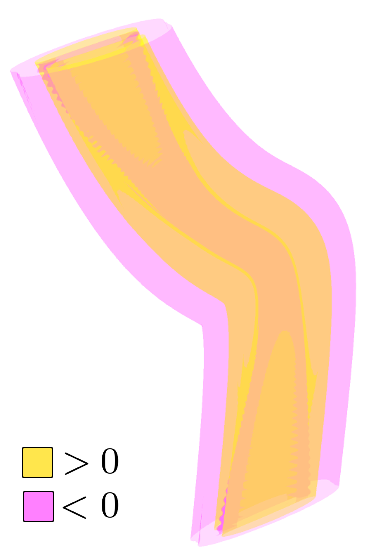}} & 
											\raisebox{-.5\height}{\includegraphics[width=0.09\textwidth]{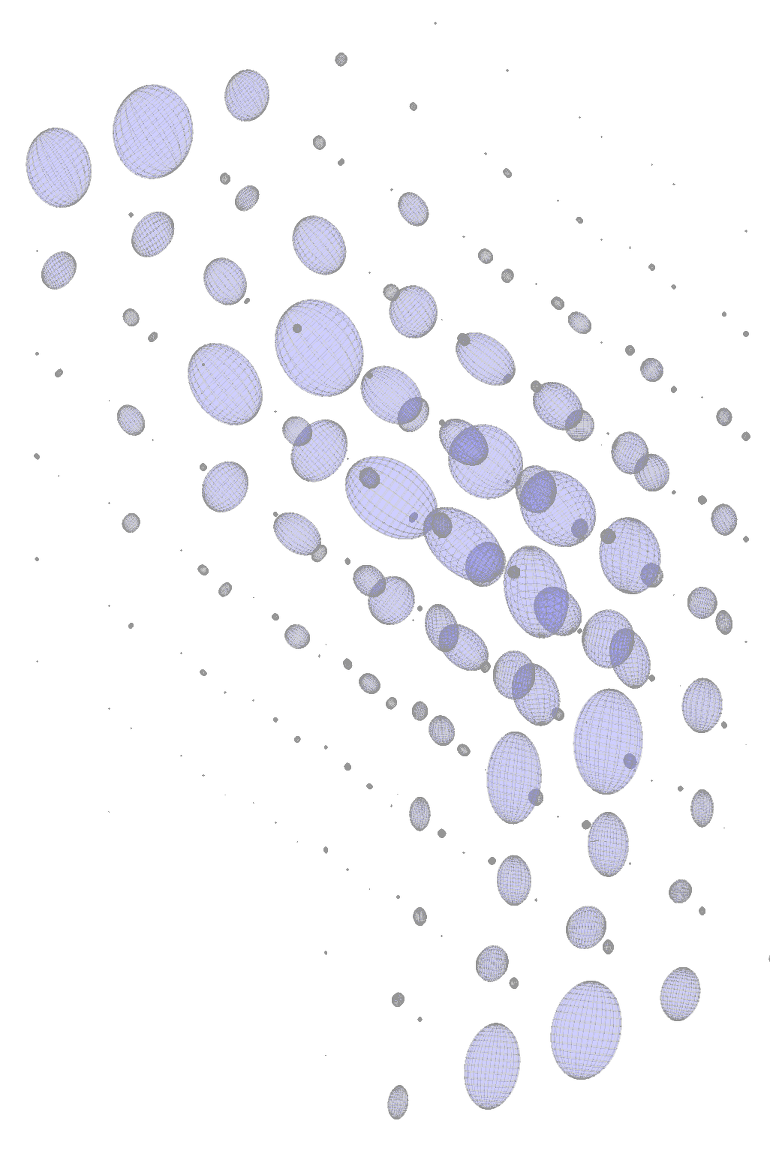}}
										\end{tabular}
	 								   } &
	\raisebox{-.5\height}{ \includegraphics[scale=0.15]{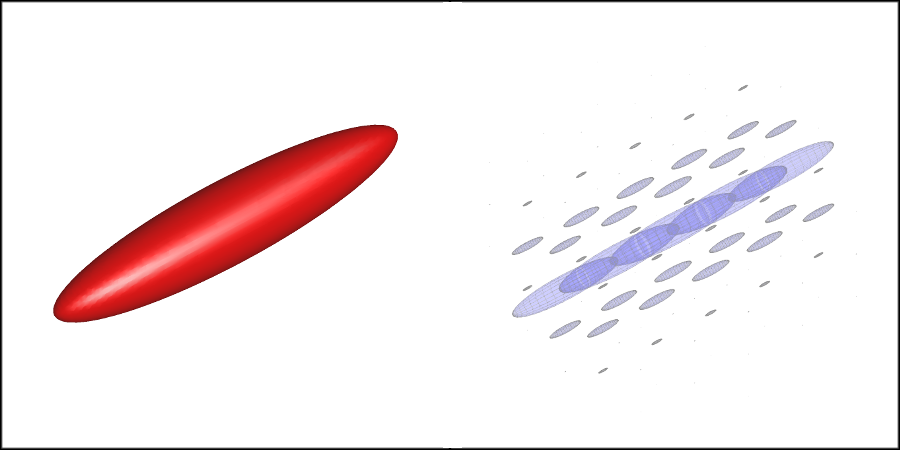} } & 
	\multirow{3}{*}[0.5em]{ \includegraphics[scale=0.11]{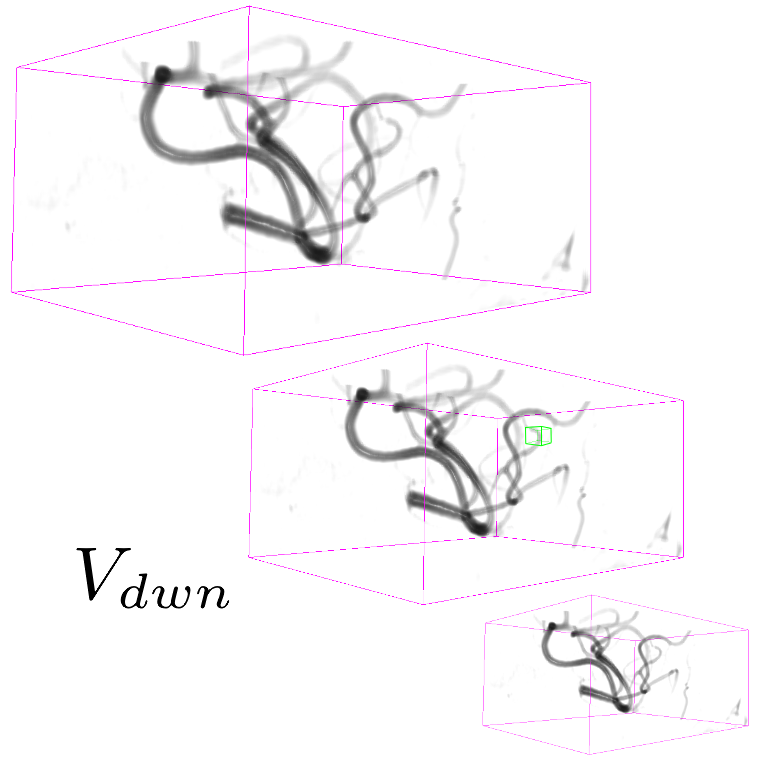} } & 
	\multirow{3}{*}[2.2em]{ 
										\begin{tabular}{|@{}c@{}|}
											\multicolumn{1}{c}{\tiny{TF\textsubscript{s}\textsuperscript{(\textit{b})}}} \\
											\hline
											\raisebox{-.5\height}{\includegraphics[width=0.275\textwidth]{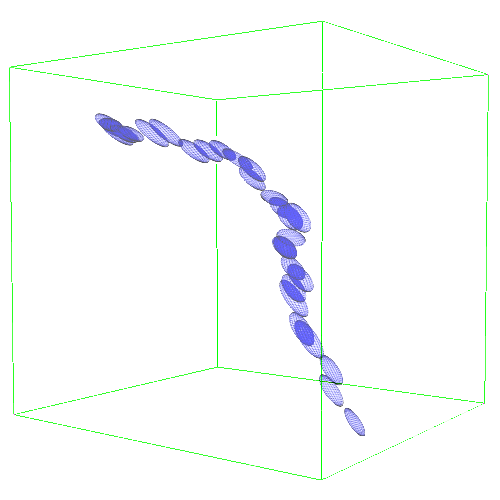}} \\ 
											\hline
										\end{tabular}
							 } \\
	&	 
	\raisebox{-.5\height}{ \includegraphics[scale=0.15]{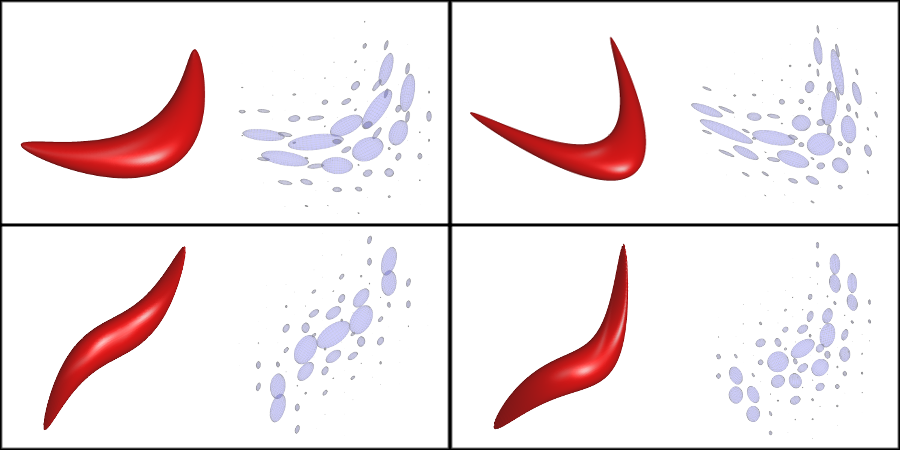} } & 
	&
	\\
	&
	\raisebox{-.5\height}{ \includegraphics[scale=0.15]{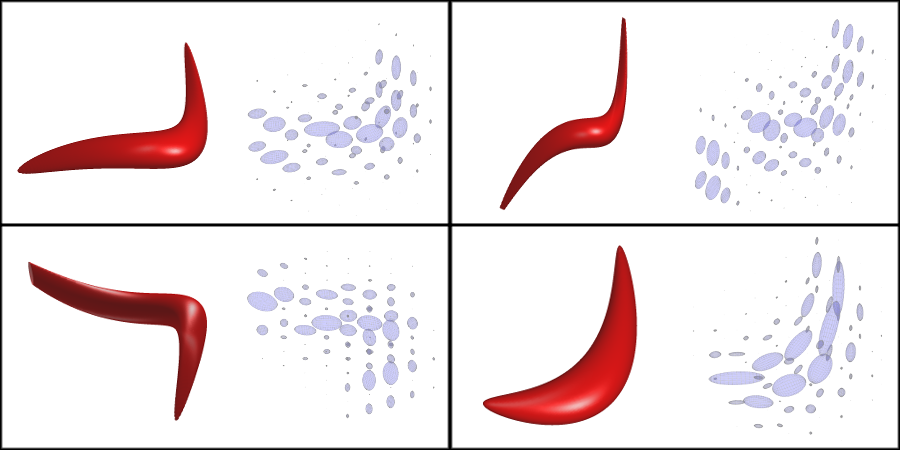} } &
	& \\
\begin{tabular}{@{}c@{}c@{}c@{}} \tiny{$K_{(\Gamma)}$} & ~~~~~~~ & \tiny{$T_{(\Gamma)}$} \end{tabular} & \tiny{DFK} & & 
\end{tabular}
\caption{{SLoGS filterbank: definition of a Dictionary of Filtering Kernels and synthesis of the Tensor Field within the overlap-add block $b$ at the given scale $s$.}}
\label{fig1}
\end{figure}
\subsection{SLoGS Curvilinear Filterbank}
With the aim of enhancing the connectivity of fragmented, furcating and tortuous vessels, we propose a multi-resolution analysis/synthesis filterbank of Steerable Laplacian of Gaussian Swirls, whose elongated and curvilinear Gaussian kernels recover a smooth, connected and orientation aware vesselness map with local maxima at vessels' mid-line.
\subsubsection{Steerable Laplacian of Gaussian Swirls (SLoGS).}
Similarly to \cite{Annunziata2015} and without losing generality, given an image \small $V: \mathbb{R}^{3} \rightarrow \mathbb{R}$, \normalsize the respective \small SLoGS \normalsize vesselness response is obtained as \small $V_{SLoGS,s} := V_{s} \ast K$, \normalsize for any given scale \small $s$ \normalsize and any predefined \small SLoGS \normalsize filtering kernel \small $K \colon \mathbb{R}^{3} \rightarrow \mathbb{R}$. \normalsize
Here we formulate and derive the \small SLoGS \normalsize filtering kernel \small $K$ \normalsize by computing the second-order directional derivative in the gradient direction of a curvilinear Gaussian trivariate function \small \mbox{$\Gamma \colon \mathbb{R}^{3} \times \mathbb{R}^{3}_{+} \times \mathbb{R}^{3} \rightarrow \mathbb{R}$}. \normalsize
The gradient direction and its perpendicular constitute the first-order gauge coordinates system \small $(\boldsymbol{\omega},\boldsymbol{\upsilon})$. \normalsize These are defined as \small $\boldsymbol{\omega} = \frac{\nabla \Gamma}{\| \nabla \Gamma \|} $, \normalsize and \small $\boldsymbol{\upsilon} = \boldsymbol{\omega}_{\perp}$, \normalsize with the spatial gradient \small $\nabla$. \normalsize The function \small $\Gamma$ \normalsize has the form
\scriptsize
\begin{equation}
\mathlarger{\Gamma\left( \mathbf{x},\boldsymbol{\sigma} ,\mathbf{c}\right)} = \frac{1}{\sqrt{2\pi\sigma_{1}^{2}}} e^{-\frac{x_{1}^{2}}{2\sigma_{1}^{2}}} \frac{1}{\sqrt{2\pi\sigma_{2}^{2}}} e^{-\frac{\left(x_{2} + c_{0}x_{1} + c_{1}x_{1}^{2}\right)^{2}}{2\sigma_{2}^{2}}} \frac{1}{\sqrt{2\pi\sigma_{3}^{2}}} e^{-\frac{\left(x_{3} + c_{2}x_{1}^{3}\right)^{2}}{2\sigma_{3}^{2}}} \enspace ,
\label{eq1}
\end{equation}
\normalsize
where \small $\mathbf{x} = x_{1}\underline{i} + x_{2}\underline{j} + x_{3}\underline{k}$, \normalsize with \small $\{\underline{i},\underline{j},\underline{k}\}$ \normalsize the Euclidean image reference system, \small $\boldsymbol{\sigma}$ \normalsize modulates the elongation and the cross-sectional profiles of the Gaussian distribution, and the curvilinear factor \small $\mathbf{c}$ \normalsize accounts for planar asymmetry and two levels of curvilinear properties (i.e. bending and tilting orthogonally to the elongation of the distribution) by means of quadratic- and cubic-wise bending of the support, respectively. For any \small $\boldsymbol{\sigma}$ \normalsize and \small $\mathbf{c}$, $\Gamma\left( \mathbf{x},\boldsymbol{\sigma} ,\mathbf{c}\right)$ \normalsize represents therefore the smooth impulse response of the Gaussian kernel. By operating a directional derivative on \small $\Gamma$ \normalsize along \small $\boldsymbol{\omega}$, \normalsize i.e. \small $\mathcal{D}_{\boldsymbol{\omega}}$, \normalsize we obtain the \small SLoGS \normalsize filtering kernel \small $K$ \normalsize as
\scriptsize
\begin{equation}
\mathlarger{K} = \mathcal{D}_{\boldsymbol{\omega}} \left[ \mathcal{D}_{\boldsymbol{\omega}} \Gamma \right] = \mathcal{D}_{\boldsymbol{\omega}} \left[ \boldsymbol{\omega}^{t} \nabla \Gamma \right] \triangleq \boldsymbol{\omega}^{t} H\left(\Gamma\right) \boldsymbol{\omega} , ~  \text{\small where} ~ H\left(\Gamma\right) = 
\begin{pmatrix}
 						  \Gamma_{\underline{i}\underline{i}} & \Gamma_{\underline{i}\underline{j}} & \Gamma_{\underline{i}\underline{k}} \\
 						  \Gamma_{\underline{j}\underline{i}} & \Gamma_{\underline{j}\underline{j}} & \Gamma_{\underline{j}\underline{k}} \\
 						  \Gamma_{\underline{k}\underline{i}} & \Gamma_{\underline{k}\underline{j}} & \Gamma_{\underline{k}\underline{k}}
\end{pmatrix}
\label{eq2}
\end{equation}
\normalsize
is the Hessian matrix of the Gaussian kernel. Given that \small $\Gamma$ \normalsize is twice continuously differentiable, \small $H(\Gamma)$ \normalsize is well defined. Since \small $H(\Gamma)$ \normalsize is symmetric, an orthogonal matrix \small $Q$ \normalsize exists, so that \small $H(\Gamma)$ \normalsize can be diagonalized as \small $H(\Gamma) = Q\Lambda Q^{-1}$. \normalsize The eigenvectors \small $\underline{q}_{l}$ \normalsize form the columns of \small $Q$, \normalsize whereas the eigenvalues \small $\lambda_{l}$, \normalsize with \small $l=1,2,3$, \normalsize constitute the diagonal elements of \small $\Lambda$, \normalsize so that \small $\Lambda_{ll} = \lambda_{l}$ \normalsize and \small $\| \lambda_{1} \| \leq \| \lambda_{2} \| \leq \| \lambda_{3} \|$. \normalsize Given a point \small $\mathbf{x}$, $K(\mathbf{x})$ \normalsize can be reformulated as \small $K(\mathbf{x}) = \omega^{t} \left(Q \Lambda Q^{-1}\right) \omega$. \normalsize Geometrically, the columns of \small $Q$ \normalsize represent a rotated orthonormal basis in \small $\mathbb{R}^{3}$ \normalsize relative to the image reference system so that \small \mbox{$\underline{q}_{l}$} \normalsize are aligned to the principal directions of \small $\Gamma$ \normalsize at any given point \small $\mathbf{x}$. \normalsize The diagonal matrix \small $\Lambda$ \normalsize characterizes the topology of the hypersurface in the neighbourhood of \small $\mathbf{x}$ \normalsize (e.g. flat area, ridge, valley or saddle point in 2D) and modulates accordingly the variation of slopes, being the eigenvalues \small $\lambda_{l}$ \normalsize the second-order derivatives along the principal directions of \small $\Gamma$. \normalsize
Factorizing \small $K(\mathbf{x})$, \normalsize we obtain: \small \mbox{$K(\mathbf{x}) = (\boldsymbol{\omega}^{t} Q) \Lambda (Q^{-1} \boldsymbol{\omega})$}, \normalsize so that the gradient direction \small $\boldsymbol{\omega}$ \normalsize is mapped onto the principal directions of \small $\Gamma$ \normalsize for any point $\mathbf{x}$. \mbox{Solving \eqref{eq2}}
\scriptsize
\begin{align}
\begin{split}
\mathlarger{K(\mathbf{x})} & = \frac{1}{\Gamma_{\underline{i}}^{2} + \Gamma_{\underline{j}}^{2} + \Gamma_{\underline{k}}^{2}} 
\begin{pmatrix}
	\Gamma_{\underline{i}} \\
	\Gamma_{\underline{j}} \\
	\Gamma_{\underline{k}}
\end{pmatrix}^t
\overbrace{
\underbrace{
\begin{pmatrix}
q_{11}&q_{21}&q_{31}\\
q_{12}&q_{22}&q_{32}\\
q_{13}&q_{23}&q_{33}
\end{pmatrix}
}_{Q}
\underbrace{
\begin{pmatrix}
\lambda_{1}&0&0\\
0&\lambda_{2}&0\\
0&0&\lambda_{3}
\end{pmatrix}
}_{\Lambda}
\underbrace{
\begin{pmatrix}
q_{11}&q_{12}&q_{13}\\
q_{21}&q_{22}&q_{23}\\
q_{31}&q_{32}&q_{33}
\end{pmatrix}
}_{Q^{-1} = Q^{t}}
}^{H(\Gamma)}
\begin{pmatrix}
	\Gamma_{\underline{i}} \\
	\Gamma_{\underline{j}} \\
	\Gamma_{\underline{k}}
\end{pmatrix} =\\
& = \sum_{l=1}^{3}\gamma_{l}\lambda_{l} =
\gamma_{1}\frac{\partial^2}{\partial\underline{q}\text{\textsubscript{1}}^{2}}\Gamma + 
\gamma_{2}\frac{\partial^2}{\partial\underline{q}\text{\textsubscript{2}}^{2}}\Gamma + 
\gamma_{3}\frac{\partial^2}{\partial\underline{q}\text{\textsubscript{3}}^{2}}\Gamma ~ \triangleq ~ \boldsymbol{\gamma}LoG(\Gamma) ~,~ \text{\small where}
\end{split}
\label{eq3}
\end{align}
\small
$\gamma_{1} = \frac{(\Gamma_{\underline{i}}q_{11} + \Gamma_{\underline{j}}q_{12} + \Gamma_{\underline{k}}q_{13})^{2}}{\Gamma_{\underline{i}}^{2} + \Gamma_{\underline{j}}^{2} + \Gamma_{\underline{k}}^{2}}$, $\gamma_{2} = \frac{(\Gamma_{\underline{i}}q_{21} + \Gamma_{\underline{j}}q_{22} + \Gamma_{\underline{k}}q_{23})^{2}}{\Gamma_{\underline{i}}^{2} + \Gamma_{\underline{j}}^{2} + \Gamma_{\underline{k}}^{2}}$, and $\gamma_{3} = \frac{(\Gamma_{\underline{i}}q_{31} + \Gamma_{\underline{j}}q_{32} + \Gamma_{\underline{k}}q_{33})^{2}}{\Gamma_{\underline{i}}^{2} + \Gamma_{\underline{j}}^{2} + \Gamma_{\underline{k}}^{2}}$
\normalsize modulate the respective components of the canonical Laplacian of Gaussian \small ($LoG$) \normalsize filter oriented along the principal directions of \small $\Gamma$. \normalsize It is clear that given any arbitrary orientation \small $\Omega$ \normalsize as an orthonormal basis similar to \small $Q$, \normalsize the proposed dictionary of filtering kernels can steer by computing the rotation transform, which maps the integral orientation basis of each Gaussian kernel \small \mbox{$\Phi_{\Gamma} = \frac{\int (\Gamma(\mathbf{x}) \cdot Q(\mathbf{x})) d\mathbf{x}}{\| \int (\Gamma(\mathbf{x}) \cdot Q(\mathbf{x})) d\mathbf{x} \|}$} \normalsize on \small $\Omega$. \normalsize
Together with the \small SLoGS \normalsize filtering kernel \small $K$, \normalsize we determine the second-moment matrix \small $T$ \normalsize associated to the filter impulse response \small $\Gamma$ \normalsize by adopting the ellipsoid model in the continuous neighborhood of \small $\mathbf{x}$. \normalsize
A symmetric tensor \small $T(\mathbf{x})$ \normalsize is derived from the eigendecomposition of \small ${H(\Gamma)}$ \normalsize as \small $T(\mathbf{x}) = Q~\Psi~Q^{-1}$, \normalsize where \small $\Psi$ \normalsize is the diagonal matrix representing the canonical unitary volume ellipsoid
\scriptsize
\begin{align}
\mathlarger{\Psi} = \mathsmaller{\left(\prod_{l=1}^{3}\psi_{l}\right)^{-\frac{1}{3}}} \left( \begin{smallmatrix}
\psi_{1}&0&0\\
0&\psi_{2}&0\\
0&0&\psi_{3}
\end{smallmatrix} \right) , ~~\text{\small being} ~~ \psi_{1} = \frac{|\lambda_{1}|}{\sqrt{| \lambda_{2}\lambda_{3} |}} ,~ \psi_{2} = \frac{|\lambda_{2}|}{| \lambda_{3} |} , ~~ \text{\small and} ~~ \psi_{3} = 1
\label{eq4}
\end{align}
\normalsize
the respective semiaxes' lengths.
Conversely from \small $H(\Gamma)$, \normalsize which is indeterminate, the tensor field \small $T$ \normalsize is a symmetric positive definite (SPD) matrix for any \small $\mathbf{x} \in \mathbb{R}^{3}$. \normalsize Here, the definition of the tensor kernel \small $T$ \normalsize in \eqref{eq4} can be further reformulated exploiting the intrinsic log-concavity of \small $\Gamma$. \normalsize By mapping \small \mbox{$\Gamma \mapsto \tilde{\Gamma} = -\log(\Gamma)$}, \normalsize a convex quadratic form is obtained, so that \small $H(\tilde{\Gamma})$ \normalsize is an SPD, as the modelled tensor \small $T$. \normalsize In either case, the manifold of tensors can be mapped into a set of 6 independent components in the Log-Euclidean space, which greatly simplifies the computation of Riemannian metrics and statistics. We refer to \cite{arsigny2006log} for a detailed methodological description. The continuous and smooth tensor field \small $T$ \normalsize inherits the steerable property. Similarly to diffusion tensor MRI, the kernel shows a preferred diffusion direction for a given energy potential, e.g. the scalar function \small $\Gamma$ \normalsize itself (\cref{fig1}). This allows to define an arbitrary dictionary of filtering kernels \small (DFK) \normalsize that embeds anisotropy and high-order directional features to scalar curvilinear templates, which enhances and locally resembles typical, smooth vessel patterns.
Together with the arbitrary \small SLoGS DFK, \normalsize we also introduce an extra pair of non-curvilinear kernels for completeness. These are the pseudo-impulsive \small $\delta LoG$, \normalsize an isotropic derivative filter given by the Laplacian of Gaussian of \small $\Gamma_{\delta}(\mathbf{x},\boldsymbol{\sigma},\mathbf{c} = \mathbf{0})$, \normalsize representing a Dirac delta function for \small $\boldsymbol{\sigma}\rightarrow 0$. \normalsize Also, the uniformly flat \small $\nu LoG$ \normalsize is another isotropic degenerate case, where the Laplacian of Gaussian derives from \small $\Gamma_{\nu}(\mathbf{x},\boldsymbol{\sigma},\mathbf{c} = \mathbf{0})$, \normalsize which is assumed to be a uniform, constant-value kernel for \small $\boldsymbol{\sigma}\rightarrow \infty$. \normalsize The purpose of introducing the extra kernels is to better contrast regions that most likely relate to vessel boundaries and to image background, respectively. Although \small $\delta LoG$ \normalsize and \small $\nu LoG$ \normalsize have singularities, ideally they represent isotropic degenerate kernels. Therefore we associate pure isotropic tensors for any given \small $\mathbf{x} \in \mathbb{R}^{3}$, \normalsize so that \small $T_{\delta LoG}(\mathbf{x}) = T_{\nu LoG}(\mathbf{x}) = I_{3} $ (Identity). \normalsize The respective directional kernel bases \small $\Phi(\delta LoG) = \Phi(\nu LoG)$ \normalsize are undetermined.
\subsubsection{Connected Vesselness Map and the Tensor Field.}
The idea is to convolve finite impulse response \small SLoGS \normalsize with the discrete vascular image in a scale- and rotation-invariant framework, to obtain simultaneously the connected vesselness maps and the associated tensor field. For simplicity, the filtering steps will be presented for a generic scale \small $s$. \normalsize Scale-invariance is achieved by keeping the size of the small compact-support \small SLoGS \normalsize fixed, while the size of the vascular image \small $V$ \normalsize varies accordingly with the multi-resolution pyramid. Also, different \small $\boldsymbol{\sigma}$ \normalsize will produce \small SLoGS \normalsize kernels with different spatial band-pass frequencies. \small $V$ \normalsize is down-sampled at the arbitrary scale \small $s$ \normalsize as proposed in \cite{cardoso2015scale} to obtain \small $V_{dwn}$. \normalsize An early saliency map of tubular structures \small $V_{tube}$ \normalsize is then determined as
\scriptsize
\begin{align}
V_{tube} = \sum_{\Omega} V_{tube}^{(\Omega)}~ , ~~ \text{\small where} ~~ V_{tube}^{(\Omega)} = \max \left( 0, V_{dwn} \ast K_{tube}^{(\Omega)} \right).
\label{eq5}
\end{align}
\normalsize
\small $K_{tube}$ \normalsize is derived from the discretized tubular kernel \small $\Gamma_{tube}(\mathbf{x}, \sigma_{1}>\sigma_{2}=\sigma_{3},\mathbf{c} = \mathbf{0})$ \normalsize (\cref{fig1}), whereas \small $\Omega$ \normalsize is defined as a group of orthonormal basis in \small $\mathbb{R}^{3}$, \normalsize using an icosphere at arbitrary subdivision level \small $n$ \normalsize to determine the orientation sampling in 3D.
\small $V_{tube}$ \normalsize is meant to provide an initial, coarse, although highly-sensitive set of saliency features in \small $V_{dwn}$: \normalsize the vessel \textit{spatial locations} and \textit{orientations}. The identification of such features has two advantages; firstly it restricts the problem of the rotation-invariant filtering to an optimal complexity in 3D avoiding unnecessary convolutions; secondly it allows to use a locally selective overlap-add \small (OLA) \normalsize \cite{oppenheim2010discrete} for the analysis/synthesis filtering. In detail, vessel spatial locations are mapped as voxel seeds \small $\tilde{S}$, \normalsize and the associated set of orientations \small $\Theta$ \normalsize forms a group of orthonormal basis in \small $\mathbb{R}^{3}$. \normalsize We define \small $\tilde{S}$ \normalsize as
\scriptsize
\begin{equation}
\tilde{S} = \text{div} \left( \nabla V_{tube} \right) _{<0} ~ \wedge ~ \lambda_{1,2,3}^{V_{tube}} < 0 ~ \wedge ~ V_{tube} \geq Q_{p}(V_{tube}^{+}) \enspace ,
\label{eq6}
\end{equation}
\normalsize
where \small $\text{div} \left( \nabla V_{tube} \right)$ \normalsize is the divergence of \small $V_{tube}$'s \normalsize gradient vector field, \small $\lambda_{1:3}^{V_{tube}}$ \normalsize are the eigenvalue maps derived from the voxel-wise eigendecomposition of \small $H(V_{tube})$, \normalsize and \small $Q_{p}(V_{tube}^{+})$ \normalsize is the \textit{p\textsuperscript{th}} quantile of the positive \small $V_{tube}$ \normalsize samples' pool. With \small $\tilde{S}$, \normalsize the orientations \small $\Theta$ \normalsize are automatically determined as the set of eigenvectors associated to \small $\lambda_{1:3}^{V_{tube}}$. \normalsize The greater the intensity threshold \small $Q_{p}(V_{tube}^{+})$, \normalsize the greater the image noise-floor rejection, the lower the number of seeds and the fewer the details extracted from \small $V_{tube}$. \normalsize Also, the cardinality of \small $\tilde{S}$ \normalsize and \small $\Theta$ \normalsize is a trade-off for the convolutional complexity in each \small OLA \normalsize filtering step. The analysis/synthesis filtering can be embedded in a fully parallel \small OLA, \normalsize by considering an overlapping grid of 3D cubic blocks spanning the domain of \small $V_{dwn}$, \normalsize and by processing each block \small $b$ \normalsize so that at least one seed exists within it.
The integral connected vesselness map \small $\text{\textit{CVM}}_{s}^{(b)}$, \normalsize for each block \small $b$ \normalsize at any given scale \small $s$, \normalsize has the form
\scriptsize
\begin{equation}
\text{\textit{CVM}}_{s}^{(b)} = \sum_{K \in DFK} \, \sum_{\theta \in \Theta^{(b)}} V_{\mathcal{S}}^{(b,K,\theta)}, ~~\text{\small where} ~~ V_{\mathcal{S}}^{(b,K,\theta)} = \max \left(0, \left( V_{dwn}^{(b)} \cdot \mathcal{H} \right) \ast K^{(\theta)} \right).
\label{eq7}
\end{equation}
\normalsize
Here, \small $V_{\mathcal{S}}^{(b,K,\theta)}$ \normalsize is the convolutional filter response given the considered \small SLoGS \normalsize kernel. In detail, \small $V_{dwn}^{(b)}$ \normalsize is the down-sampled image in \small $b$, $\mathcal{H}$ \normalsize is the 3D \small OLA \normalsize Hann weighting window, and \small $K^{(\theta)}$ \normalsize is the steered filtering kernel along \small $\theta \in \Theta^{(b)}$, \normalsize those being the seeds' orientations in \small $b$. \normalsize Note that in the discrete domain each voxel has a spatial indexed location \small $\mathbf{b} \in b$. \normalsize
The anisotropic tensor field \small $\text{\textit{TF}}_{s}^{(b)}$ \normalsize is synthesized and normalized in the Log-Euclidean space as the integral \mbox{\textit{weighted-sweep}} of each steered tensor patch within the block \small $b$, \normalsize and has the form
\scriptsize
\begin{equation}
\begin{smallmatrix}
\mathlarger{\text{\textit{TF}}_{s,(LE)}^{(b)}} = \dfrac{1}{W} ~~\cdot \mathlarger{\mathlarger{\sum}}_{K \in \left\lbrace \begin{smallmatrix} \scriptscriptstyle DFK,\\ \scriptscriptstyle \delta LoG,\\ \scriptscriptstyle \nu LoG \end{smallmatrix} \right\rbrace } ~ \mathlarger{\sum}_{\theta \in \Theta^{(b)}} \underbrace{ \left( \mathsmaller{\sum}_{\lfloor \mathbf{b} \rceil \subset b} \overbrace{ V_{\mathcal{S}}^{(\mathbf{b},K,\theta)} \cdot \Gamma_{(K)}^{(\theta)} \cdot \Xi }^{\text{\scriptsize weights}} \cdot \overbrace{ T^{(\theta)}_{K,(LE)} }^{\text{\scriptsize patch}} \right) }_{\text{\scriptsize within-block patch sweep}}, ~~ \text{\small so that}\\
\mathlarger{ \det{\left( \text{\textit{TF}}_{s}^{(b)} (\mathbf{b}) \right) = \mathcal{H}(\mathbf{b})} }, ~~ \text{\small and~} \, \mathlarger{W} =  \left( \sum_{K \in \left\lbrace \begin{smallmatrix} \scriptscriptstyle DFK,\\ \scriptscriptstyle \delta LoG,\\ \scriptscriptstyle \nu LoG \end{smallmatrix} \right\rbrace } ~ \sum_{\theta \in \Theta^{(b)}} \sum_{\lfloor \mathbf{b} \rceil \subset b} V_{\mathcal{S}}^{(\mathbf{b},K,\theta)} \cdot \Gamma_{(K)}^{(\theta)} \cdot \Xi \right),
\end{smallmatrix}
\label{eq8}
\end{equation}
\normalsize
where \small $W$ \normalsize is the integral normalizing weight-map accounting for all vessel, boundary and background components; \small $V_{\mathcal{S}}^{(\mathbf{b},K,\theta)}$ \normalsize is the modulating \small SLoGS \normalsize filter response at \small $\mathbf{b}$ \normalsize as in \eqref{eq7}; \small $\Gamma_{(K)}^{(\theta)}$ \normalsize is the steered Gaussian impulse response associated to the kernel \small $K \in \left\lbrace DFK, \delta Log, \nu LoG \right\rbrace$; $\Xi$ \normalsize is the Hann smoothing window in the neighbourhood \small $\lfloor \mathbf{b} \rceil$ \normalsize centred at \small $\mathbf{b}$, \normalsize and \small $T^{(\theta)}_{K,(LE)}$ \normalsize is one of the 6 components of the discrete steered tensors patch \small $T$ \normalsize in the Log-Euclidean space. Note that all 6 tensorial components are equally processed, and that the neighbourhood \small $\lfloor \mathbf{b} \rceil$ \normalsize and the \small SLoGS \normalsize tensors patch \small $T^{(\theta)}_{K,(LE)}$ \normalsize have the same size.
In \eqref{eq8}, \small $\text{\textit{TF}}_{s,(LE)}^{(b)}$ \normalsize integrates also the isotropic contributions from vessel boundaries and background to better contrast the tubular structures' anisotropy and to reduce synthetic artifacts surrounding the vessels (\cref{fig1}). In particular, \small $\text{\textit{TF}}_{s,(LE)}^{(b)}$ \normalsize is averaged with an identically null tensor patch in the Log-Euclidean space in correspondence of boundaries and background, and \small $V_{\mathcal{S}}^{(\mathbf{b},K,\theta)}|_{\{\delta LoG, \nu LoG\}}$ \normalsize is computed as in \eqref{eq7}, where the image negative of \small $V_{dwn}^{(b)}$ \normalsize is considered.
Lastly, the connected vesselness maps and the associated synthetic tensor field are reconstructed by adding adjacent overlapping blocks in the \small OLA \normalsize 3D grid for the given scale \small $s$. \normalsize
\subsubsection{Integration over Multiple Scales.}
Each scale-dependent contribution is up-sampled and cumulatively integrated with a weighted sum
\scriptsize
\begin{equation}
\begin{array}{lcl}
\text{\textit{CVM}} = \sum_{s} \frac{1}{s}\text{\textit{CVM}}_{s}, & \text{\small ~~~and,~~~} & \text{\textit{TF}}_{(LE)} = \frac{1}{\text{\textit{CVM}}} \sum_{s} \left( \frac{1}{s} \text{\textit{CVM}}_{s}\right) \cdot T_{s,(LE)}.
\end{array}
\label{eq9}
\end{equation}
\normalsize
Vesselness contributions are weighted here by the inverse of \small $s$, \normalsize emphasizing responses at spatial low-frequencies.
We further impose that the Euclidean \small $\text{\textit{TF}}$ \normalsize has unitary determinant at each image voxel; for stability, the magnitude of the tensors is decoupled from the directional and anisotropic features throughout the whole multi-scale process, since tensors' magnitude is expressed by \small $\text{\textit{CVM}}$. \normalsize Note that with the proposed method we do not aim at segmenting vessels by thresholding the resulting $\text{\textit{CVM}}$, we rather provide a measure of vessels' connectedness with maximal response at the centre of the vascular structures. 
\subsection{Vascular Tree of Geodesic Minimal Paths}
Following the concepts first introduced in \cite{Benmansour2011Aniso}, we formulate an anisotropic front propagation algorithm that combined with an acyclic connectivity paradigm joins multiple sources \small $\tilde{S} \mapsto S$ \normalsize propagating concurrently on a Riemannian speed potential \small $\mathcal{P}$. \normalsize Since we want to extract geodesic minimal paths between points, we minimize an energy functional \small $ \mathcal{U}(\mathbf{x}) = \min_{\pi} \int_{\pi} \mathcal{P} \left( \pi( \mathbf{x} ), \pi'( \mathbf{x} ) \right) d \mathbf{x}$ \normalsize for any possible path \small $\pi$ \normalsize between two generic points along its geodesic length, so that \small \mbox{$ \| \nabla \mathcal{U} (\mathbf{x}) \| = 1$}, \normalsize and \small \mbox{$\mathcal{U}( S ) = 0$}. \normalsize The solution to the Eikonal partial differential equation is given here by the anisotropic Fast Marching (\textit{aFM}) algorithm \cite{Benmansour2011Aniso}, where front waves propagate from \small $S$ \normalsize on \small $\mathcal{P}$, \normalsize with \small $\mathcal{P}\left( \pi, \pi' \right) = \sqrt{\pi'^{t} \cdot \mathcal{M} \cdot \pi' }$ \normalsize describing the infinitesimal distance along \small $\pi$, \normalsize relative to the anisotropic tensor \small $\mathcal{M}$. \normalsize In our case, \small $\mathcal{M} = \text{\textit{TF}}$, \normalsize and \small $\pi' \propto \frac{1}{\text{\textit{CVM}}}$. \normalsize Note that the anisotropic propagation is a generalised version of the isotropic propagation medium, \small $\mathcal{M} \equiv I_{3}$. \normalsize
The acyclic connectivity paradigm is run until convergence together with the \textit{aFM} to extract the vascular tree of multiple connected geodesics \small $\Pi$. \normalsize
\subsubsection{Anisotropic FM and Acyclic Connectivity Paradigm.}
Geodesic paths are determined by back-tracing \small $\mathcal{U}$ \normalsize when different regions collide. The connecting geodesic \small $\pi$ \normalsize is extracted minimizing \small $\mathcal{U}$ \normalsize at the collision grid-points. The \textit{aFM} maps, i.e. \small $\mathcal{U}$; \normalsize the \textit{Voronoi index map} \small $\mathcal{V}$, \normalsize representing the label associated to each propagating seed; and the \textit{Tag} \small $\mathcal{T}$, \normalsize representing the state of each grid-point (Front, Visited, Far), are then updated within the collided regions, so that these merge as one and the front is consistent with the unified resulting region. This is continued until all regions merge.

\textit{Initialization.~}The seeds \small $\tilde{S}$ \normalsize are aligned towards the vessels' mid-line with a constrained gradient descent, resulting in an initial set of sources \small $S$. \normalsize All 26-connected components \small $\pi_{p}^{(S)} \in S$ \normalsize initialize the \textit{aFM} maps, i.e., \small $\mathcal{U}( \pi_{p}^{(S)}) = 0$, $\mathcal{V}( \pi_{p}^{(S)} ) = p$, $\mathcal{T}( \pi_{p}^{(S)} ) = \text{\textit{Front}}$, \normalsize and constitute also the initial geodesics \small $\pi_{p}^{(S)} \rightarrow \Pi$. \normalsize

\textit{Fast Marching Step.~}The \textit{aFM} maps are updated by following an informative propagation scheme. We refer to \cite{Benmansour2011Aniso} for the \mbox{3D} \textit{aFM} step considering the 48 simplexes in the \mbox{26-neighbourhood} of the \textit{Front} grid-point with minimal \small $\mathcal{U}$. \normalsize

\begin{figure}[tb!]
\begin{tabular}{@{}c||cc@{}|@{}c@{}c@{}|@{}c@{}c@{}|@{}c@{}c@{}|c@{}}
	\multicolumn{1}{c}{\tiny{\textbf{Initialization}}} & \multicolumn{1}{c}{} &
	\multicolumn{1}{c}{\tiny{\textbf{FM Steps}}} &
	\multicolumn{6}{c}{\tiny{\textbf{Collision, Path Extraction and Update}}} &
	\multicolumn{1}{c}{\tiny{\textbf{Convergence}}} \\
	\tiny{\textit{$\tilde{S}$ Descent}} & \multicolumn{1}{c}{} &
	\tiny{\dots} &
	\multicolumn{2}{c}{\tiny{\textit{$\Pi \leftarrow p_{~i}$}}} & 
	\multicolumn{2}{c}{\tiny{\textit{$\Pi \leftarrow p_{~i+1}$}}} &
	\multicolumn{2}{c@{}|}{\tiny{\textit{$\Pi \leftarrow p_{~i+2}$}}} &
	\tiny{\textit{Stop Criterion}} \\
	\raisebox{-.5\height}{\includegraphics[width=.1\textwidth]{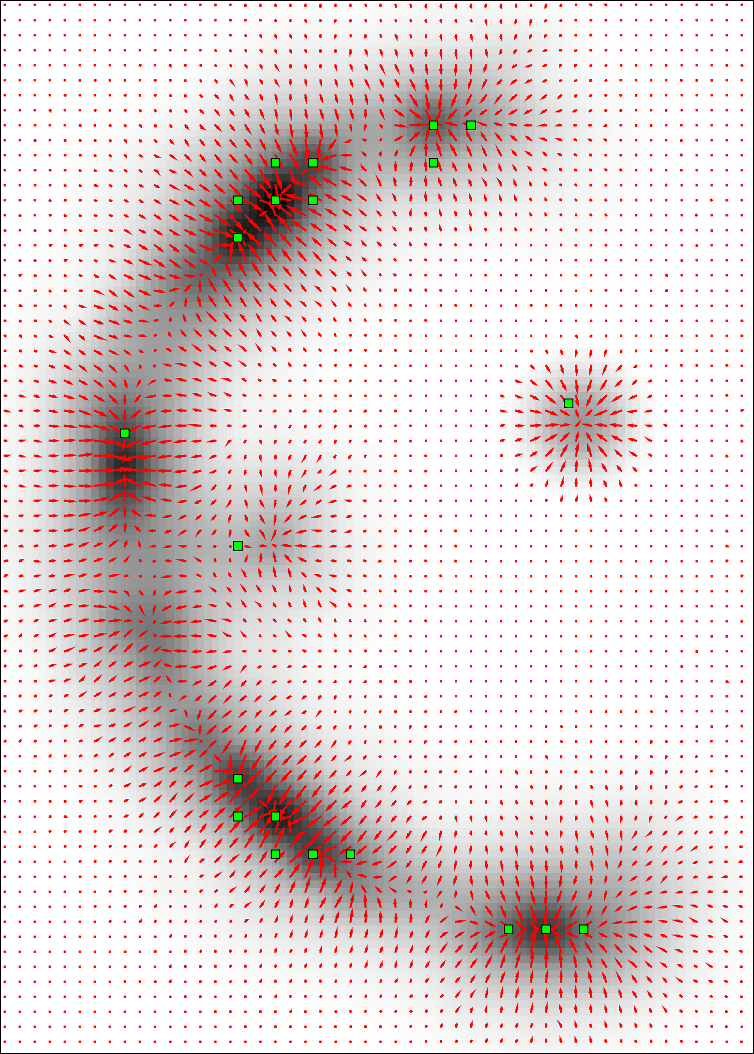}} & 
	\raisebox{-.5\height}{\begin{sideways}\centering \tiny{$\mathcal{U}~\cup$ \textit{Trials}}\end{sideways}} &
	\raisebox{-.5\height}{\includegraphics[width=.1\textwidth]{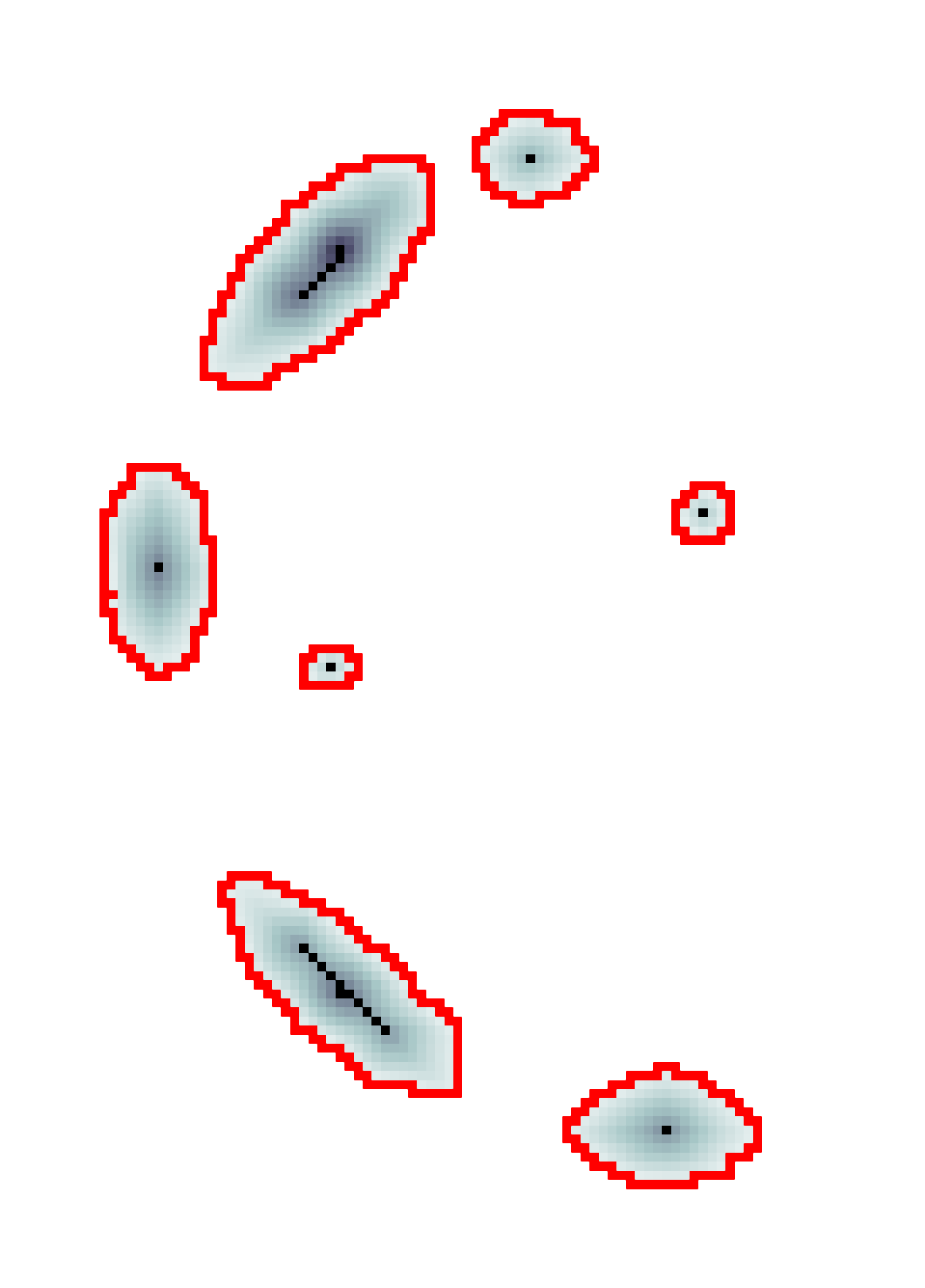}} &
	\raisebox{-.5\height}{\includegraphics[width=.1\textwidth]{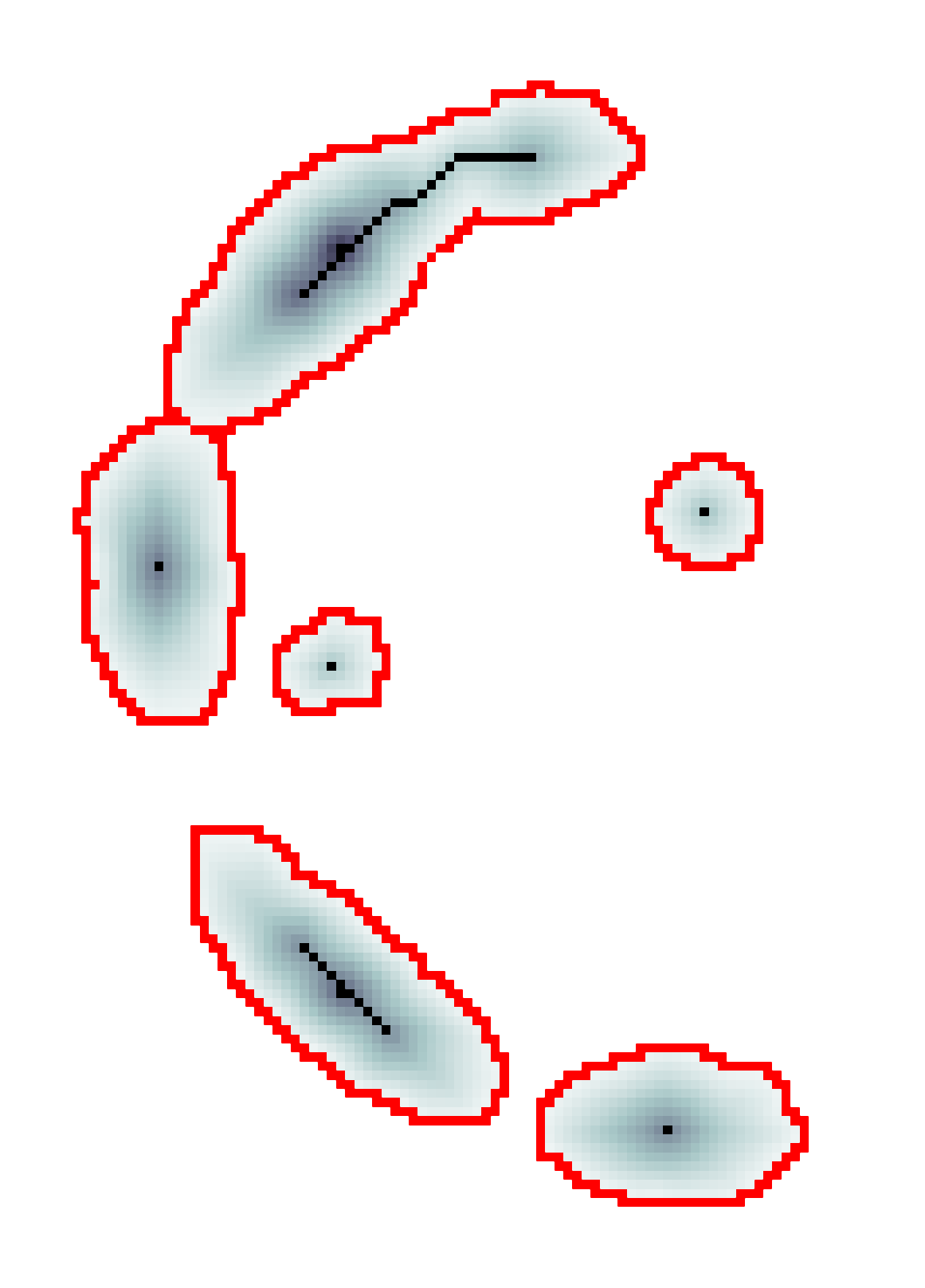}} &
	\raisebox{-.5\height}{\includegraphics[width=.1\textwidth]{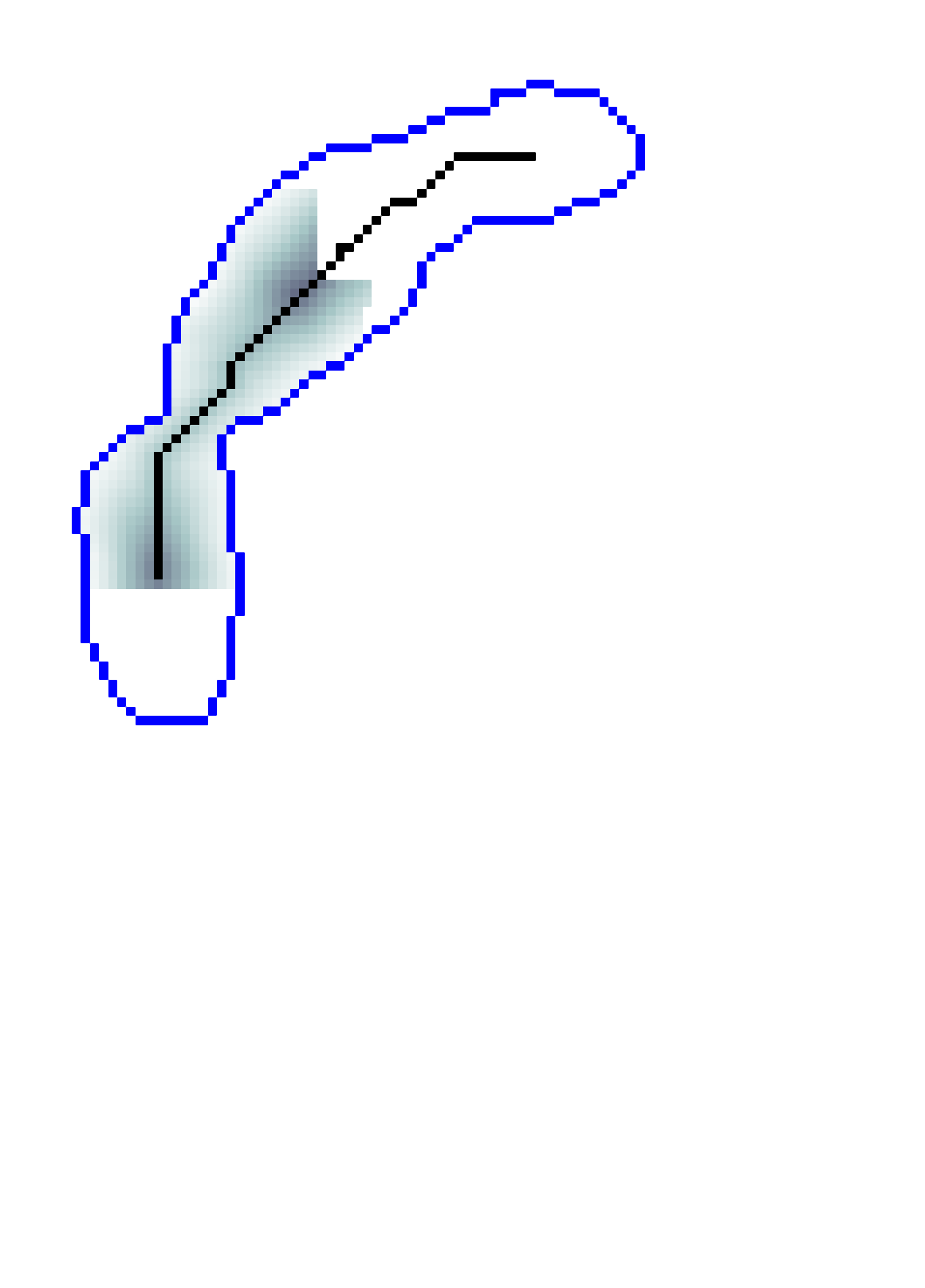}} &
	\raisebox{-.5\height}{\includegraphics[width=.1\textwidth]{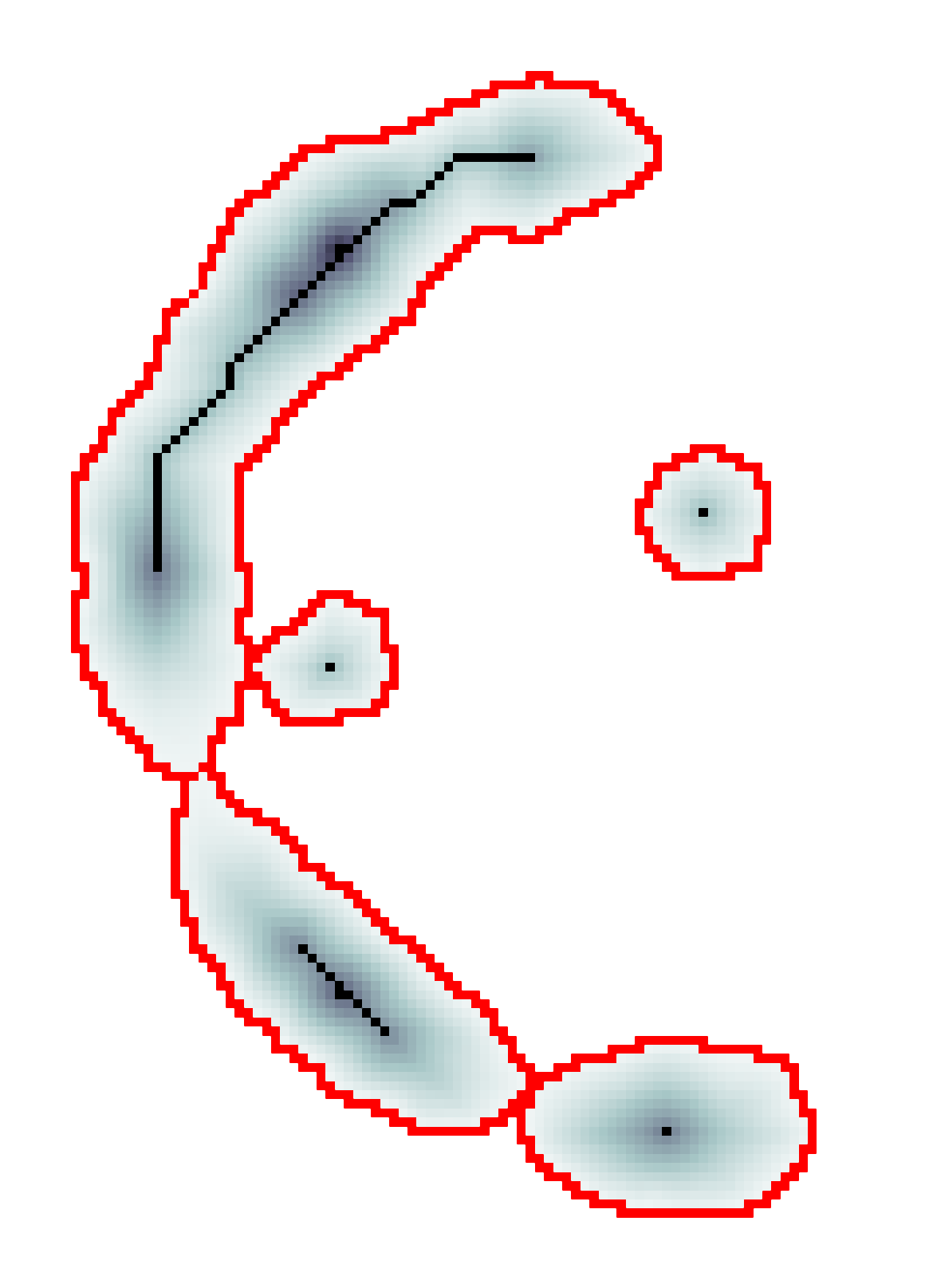}} &
	\raisebox{-.5\height}{\includegraphics[width=.1\textwidth]{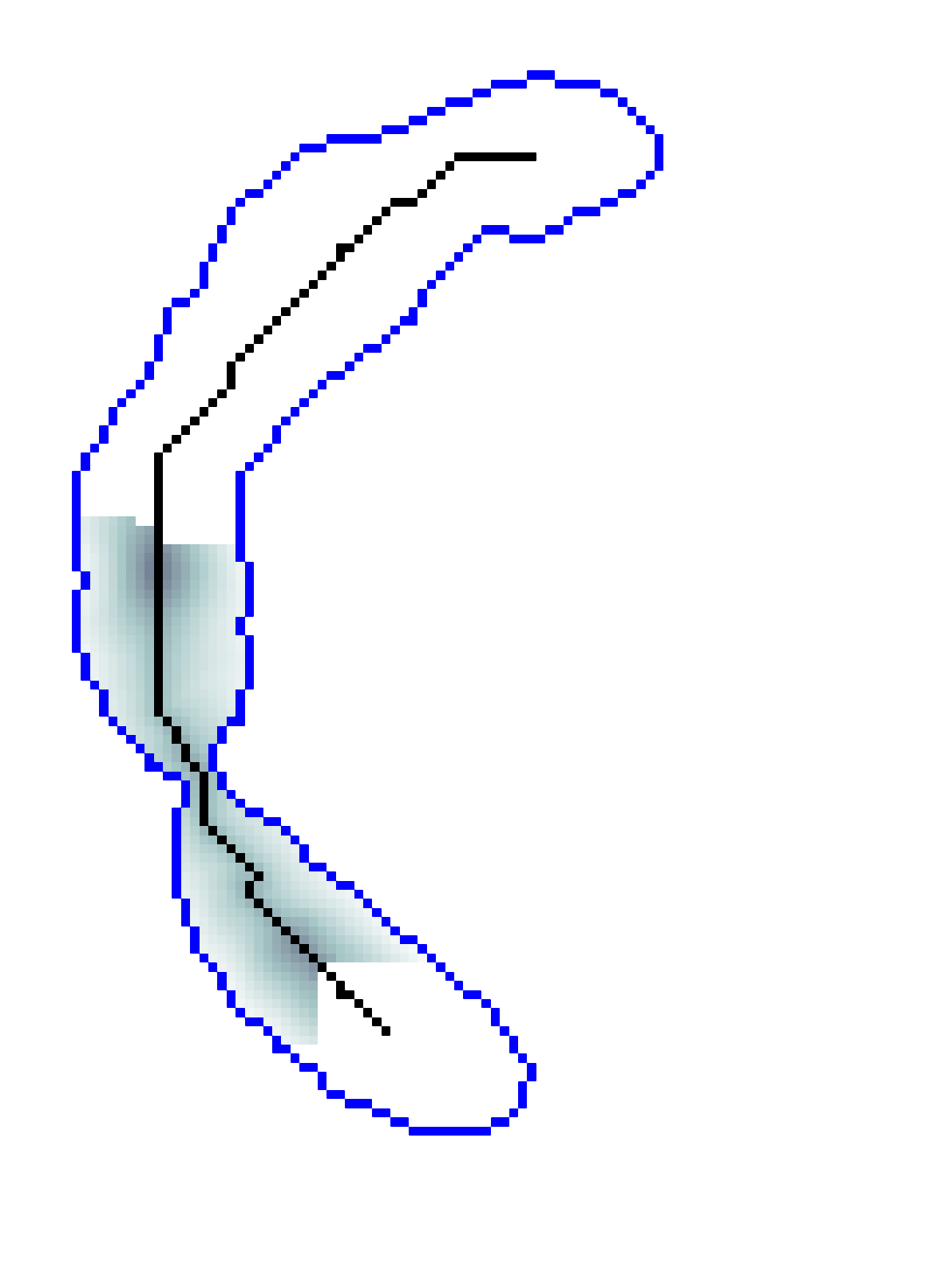}} &
	\raisebox{-.5\height}{\includegraphics[width=.1\textwidth]{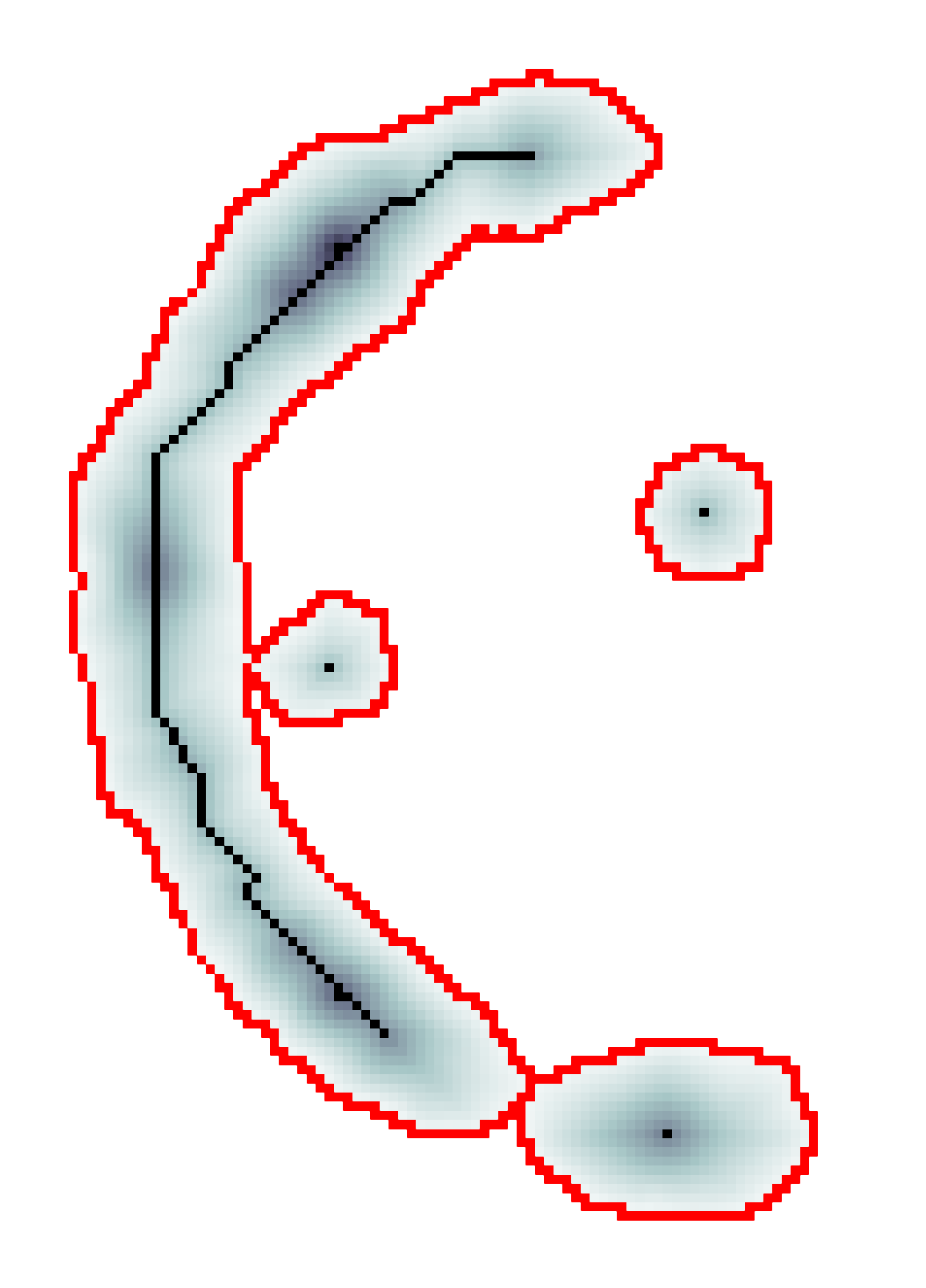}} &
	\raisebox{-.5\height}{\includegraphics[width=.1\textwidth]{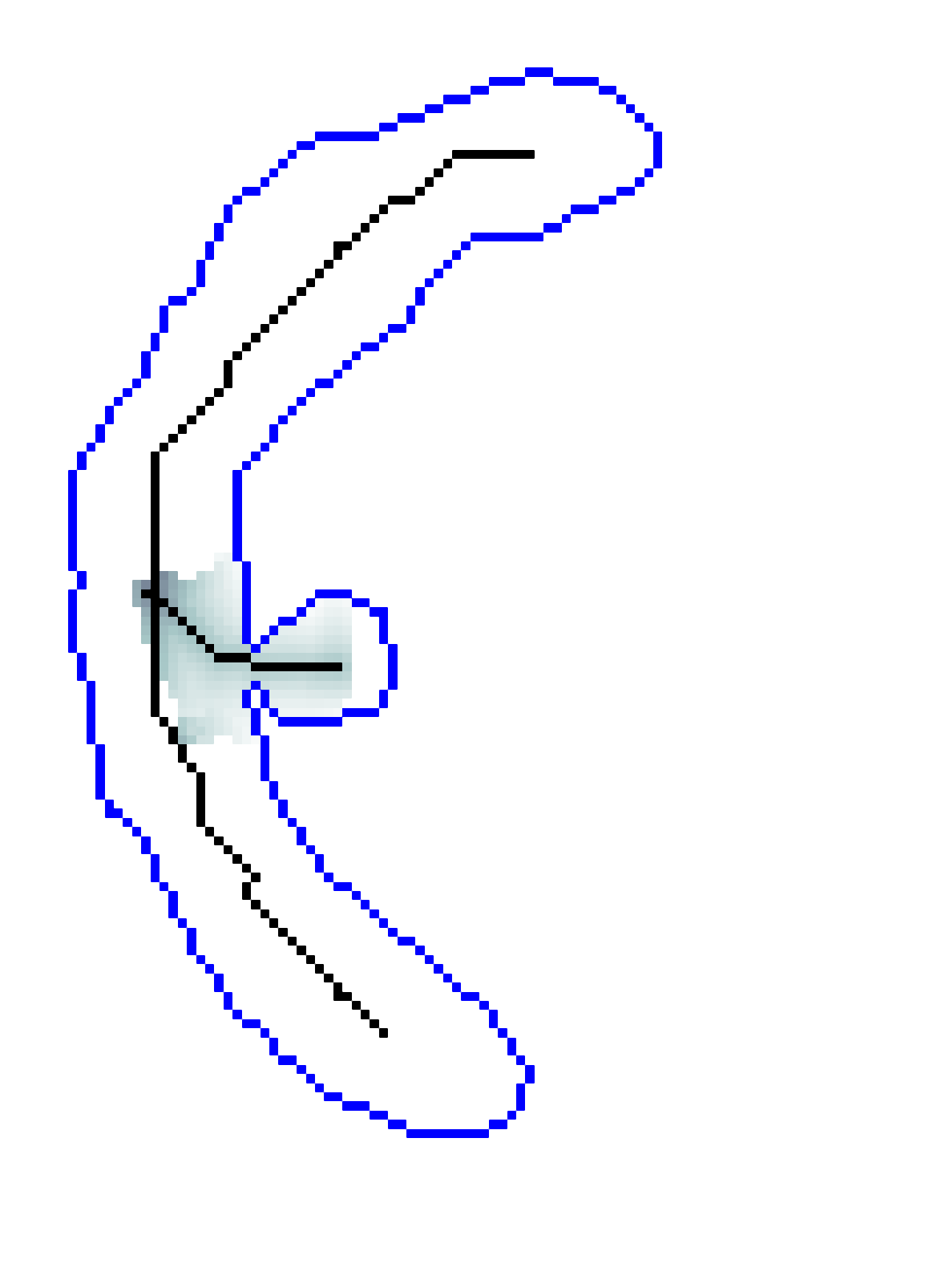}} &
	\raisebox{-.5\height}{\includegraphics[width=.1\textwidth]{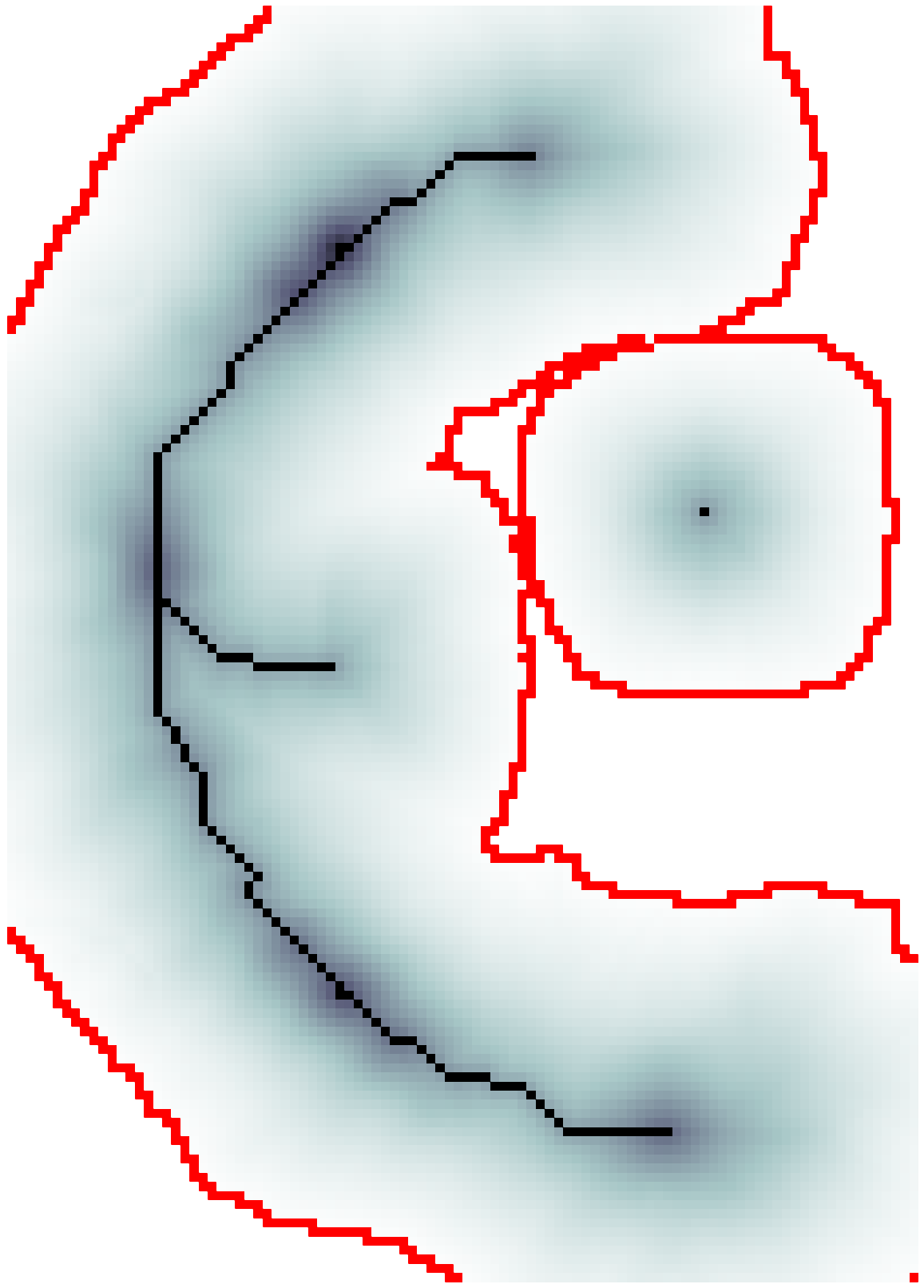}} \\
	\raisebox{-.5\height}{\includegraphics[width=.1\textwidth]{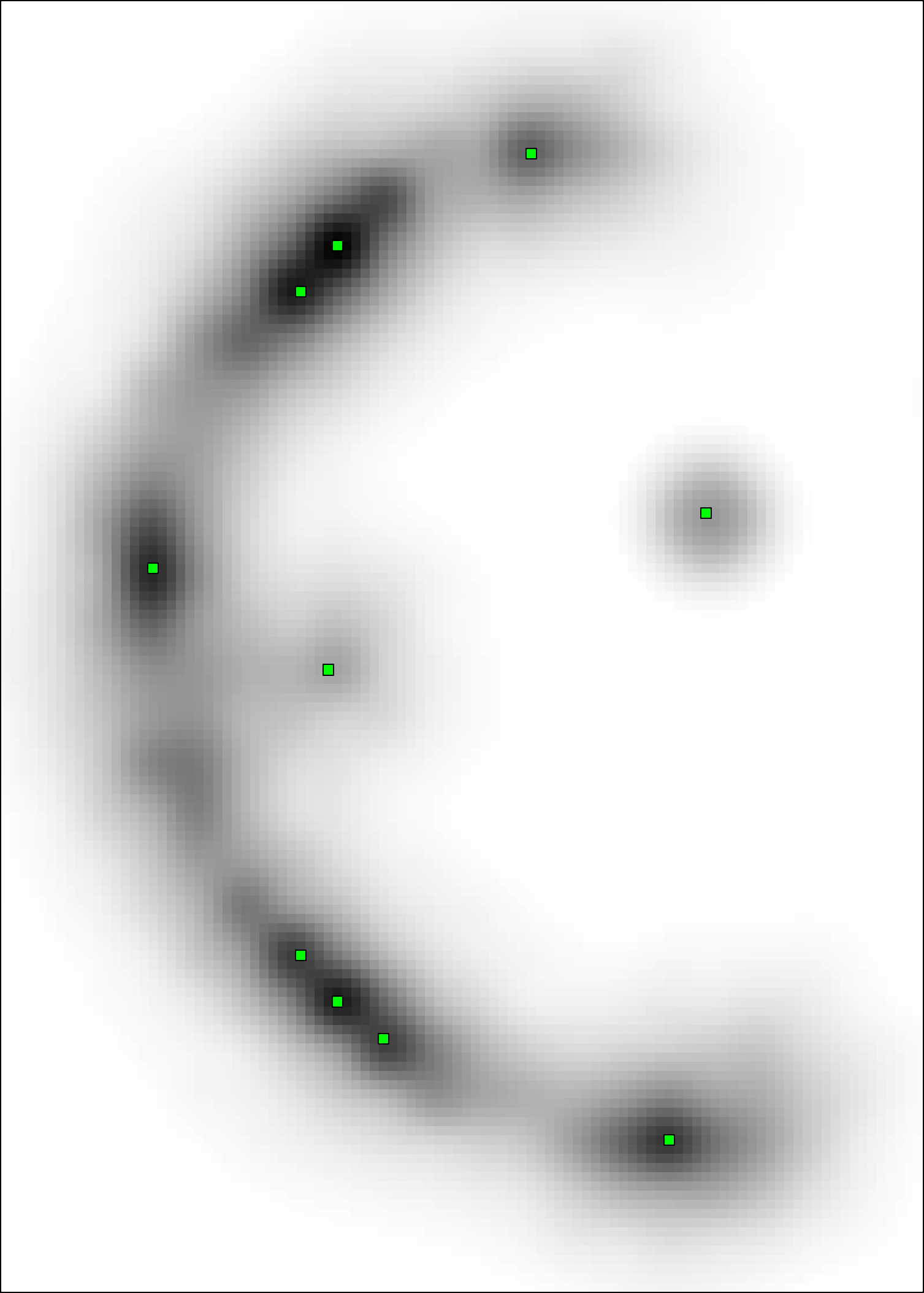}} & 
	\raisebox{-.5\height}{\begin{sideways}\centering \tiny{$\mathcal{V}$}\end{sideways}} &
	\raisebox{-.5\height}{\includegraphics[width=.1\textwidth]{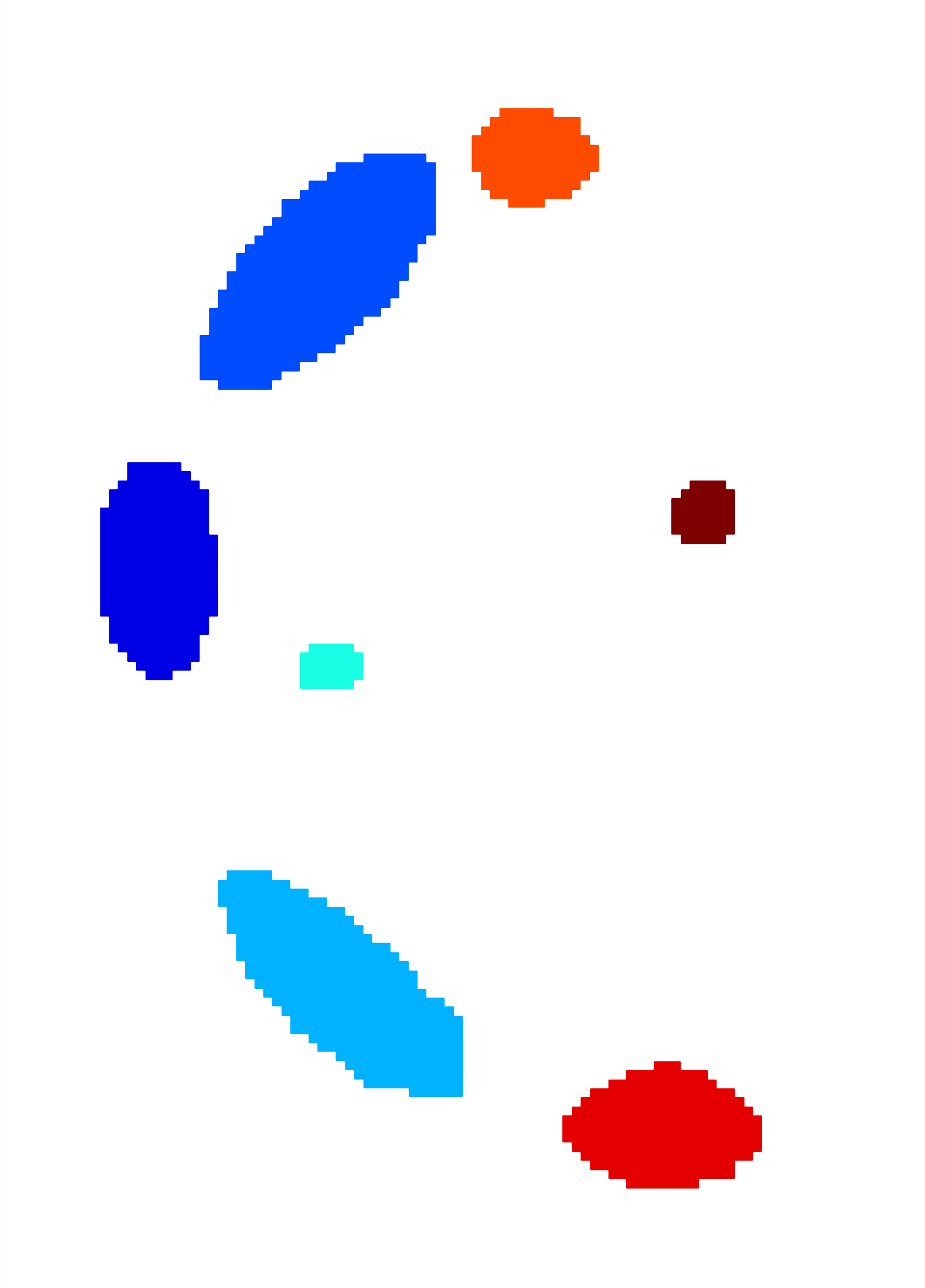}} &
	\raisebox{-.5\height}{\includegraphics[width=.1\textwidth]{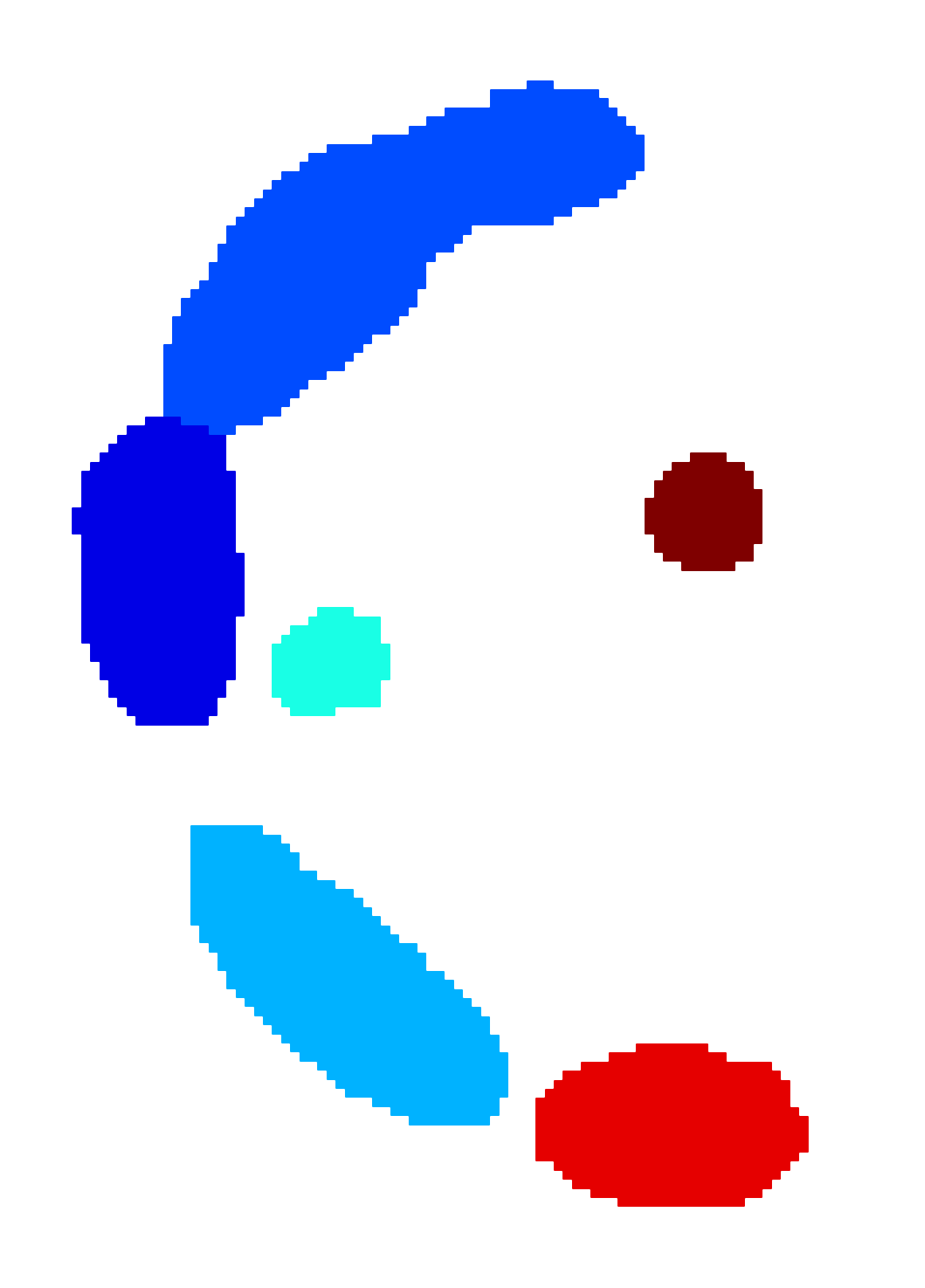}} &
	\raisebox{-.5\height}{\includegraphics[width=.1\textwidth]{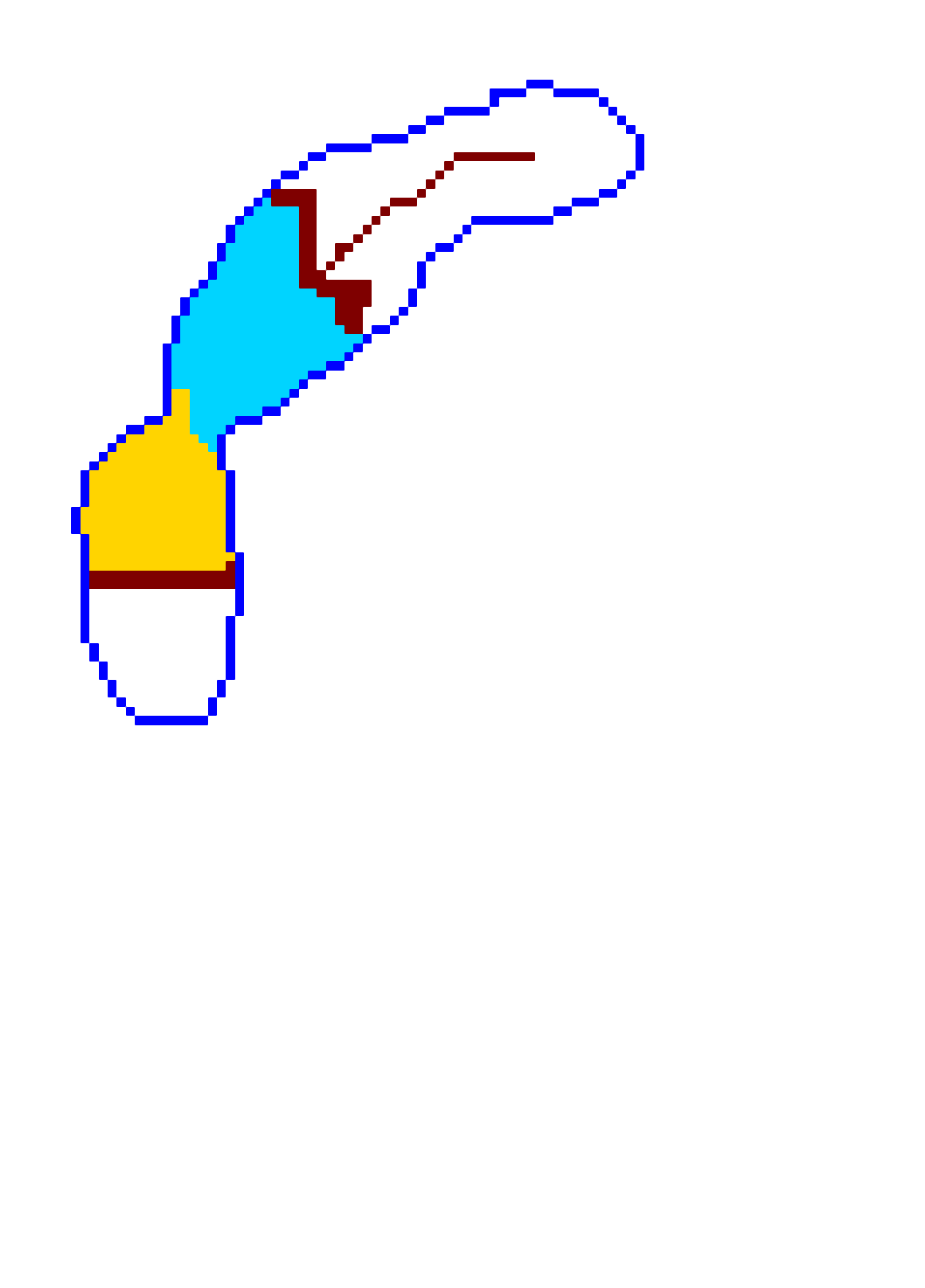}} &
	\raisebox{-.5\height}{\includegraphics[width=.1\textwidth]{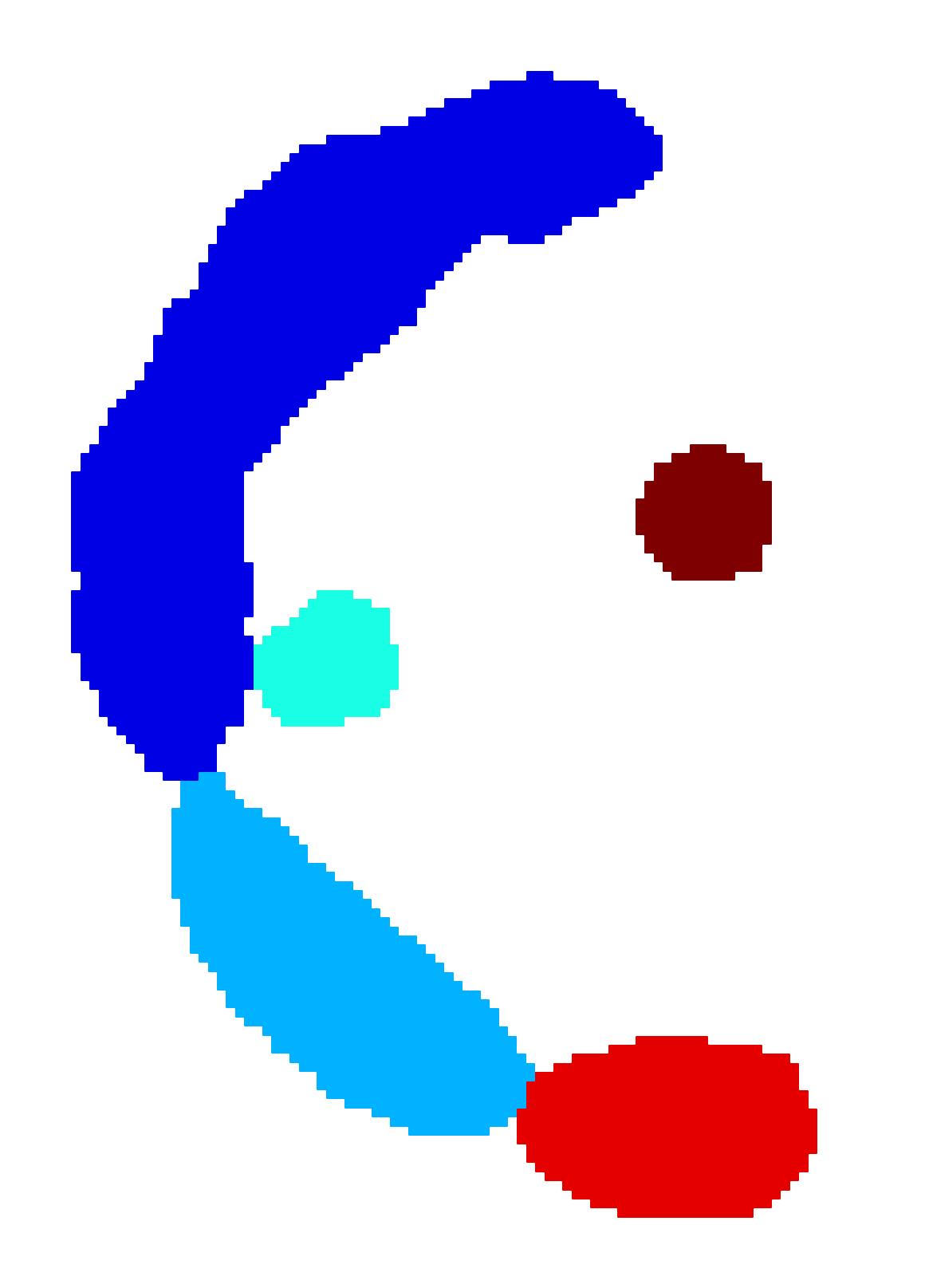}} &
	\raisebox{-.5\height}{\includegraphics[width=.1\textwidth]{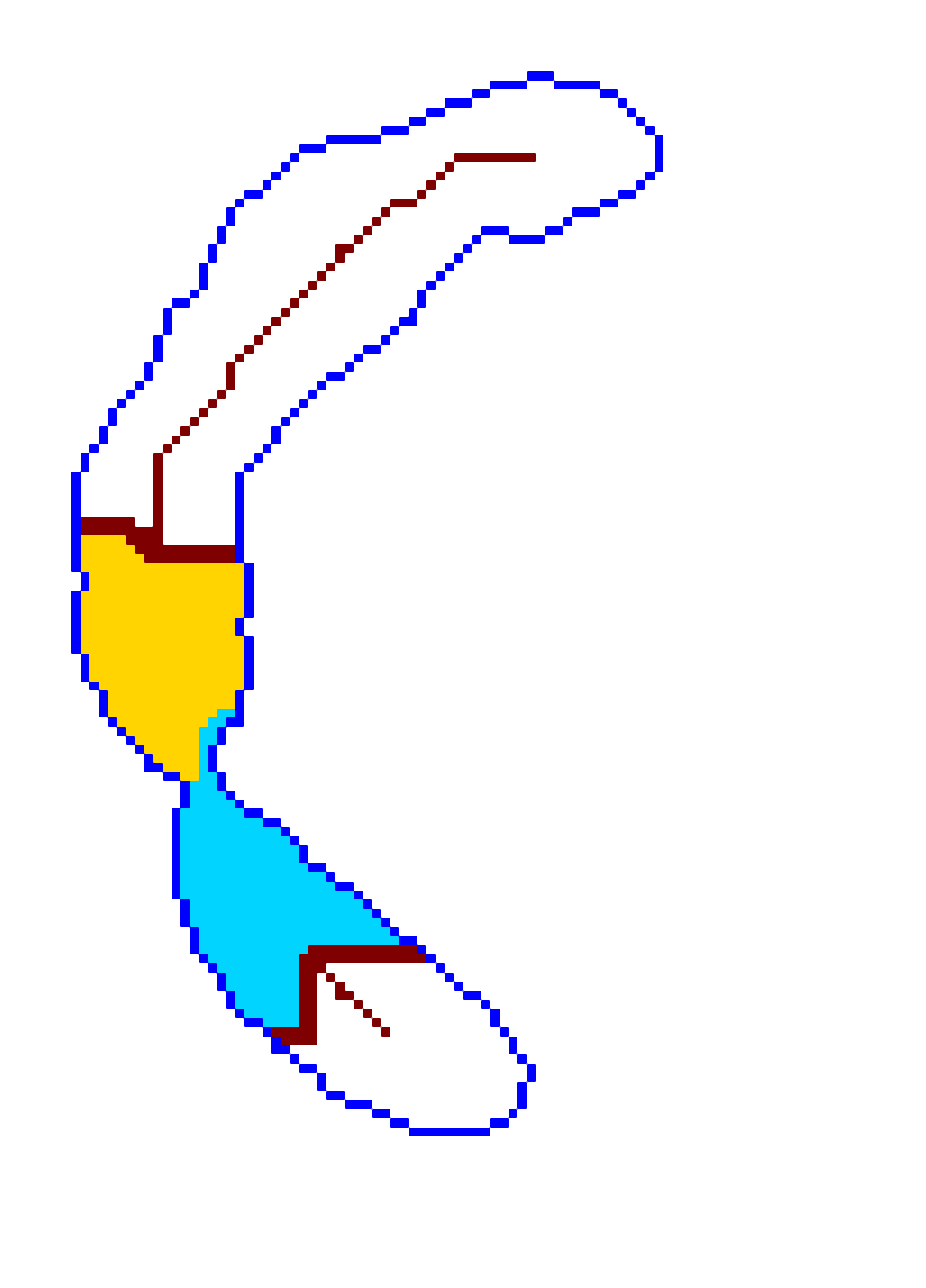}} &
	\raisebox{-.5\height}{\includegraphics[width=.1\textwidth]{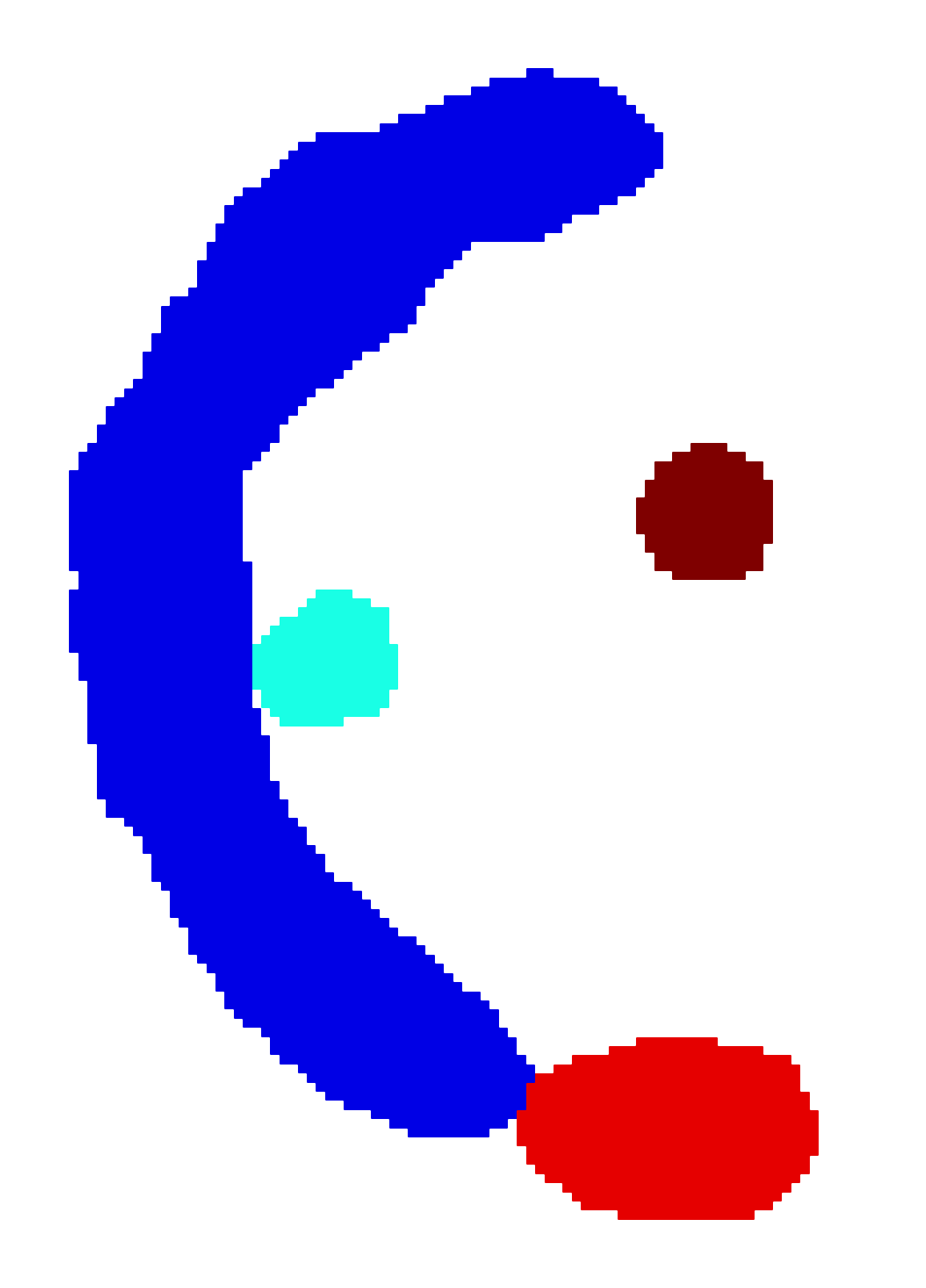}} &
	\raisebox{-.5\height}{\includegraphics[width=.1\textwidth]{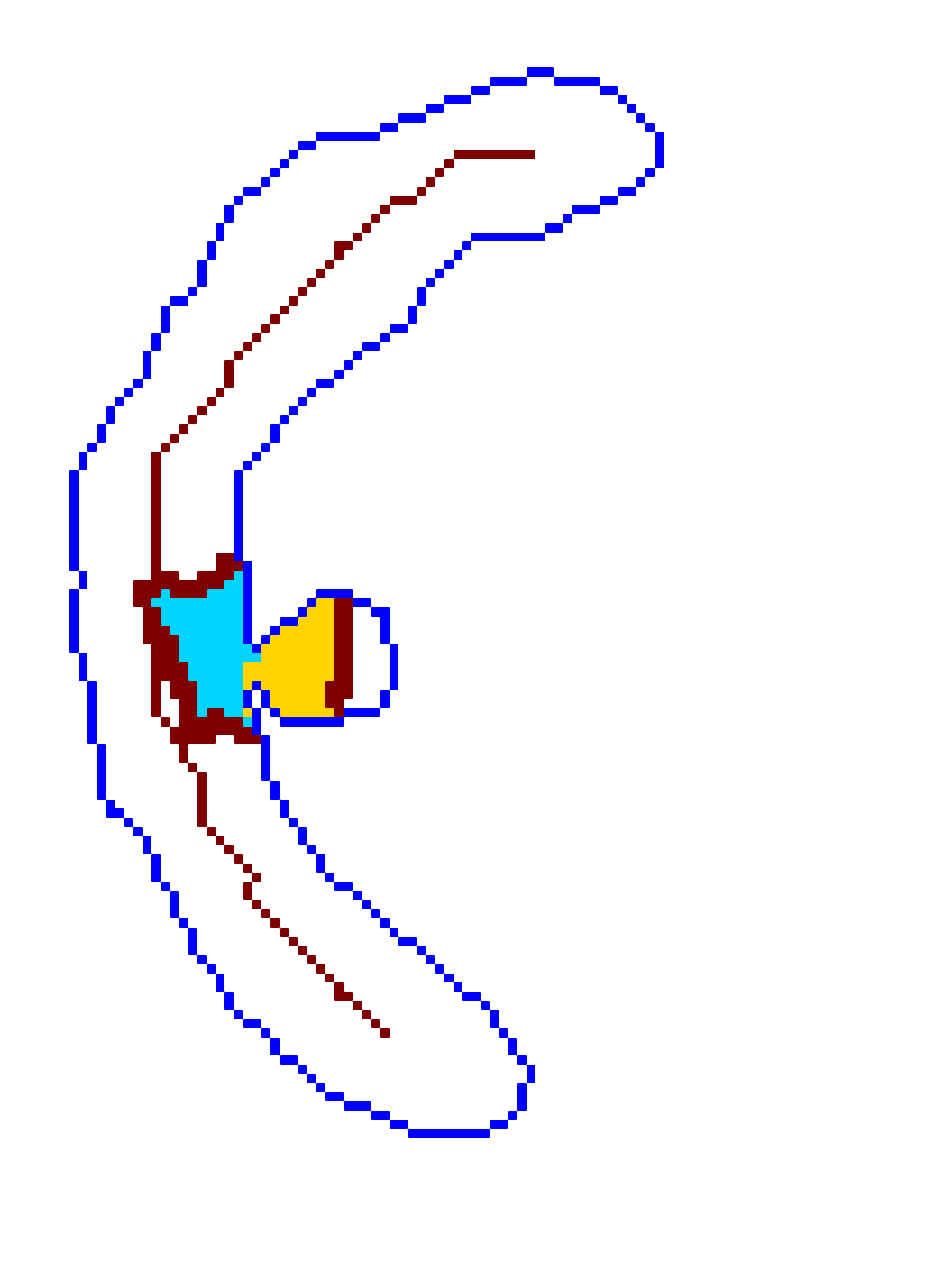}} &
	\raisebox{-.5\height}{\includegraphics[width=.1\textwidth]{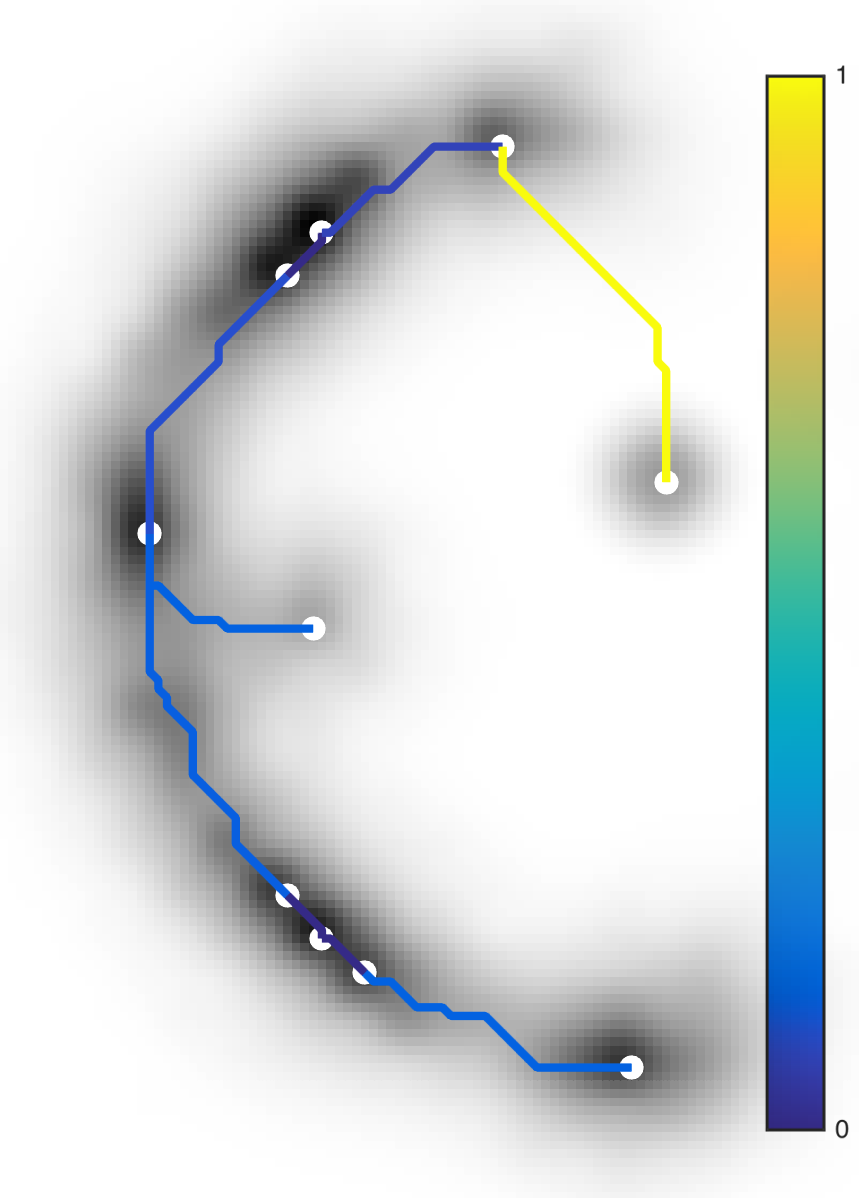}} \\
	\multicolumn{1}{c||}{\tiny{\textit{$\mathcal{P} \cup S$}}} &
	\multicolumn{3}{c}{} &
	\multicolumn{1}{c}{\tiny{$(A \cup B)^{\text{\textbf{\textit{u}}}}$}} &
	& 
	\multicolumn{1}{c}{\tiny{$(A \cup B)^{\text{\textbf{\textit{u}}}}$}} &
	&
	\multicolumn{1}{c|}{\tiny{$(A \cup B)^{\text{\textbf{\textit{u}}}}$}} &
	\tiny{\textbf{Geodesic} \textit{U}}
\end{tabular}
\caption{{Sequential acyclic connectivity paradigm on a synthetic 2D image.}}
\label{fig2}
\end{figure}

\textit{Path Extraction.~}Collision is detected when \textit{Visited} grid-points of different regions are adjacent. A connecting \small $\pi$ \normalsize is determined by linking the back-traced minimal paths from the collision grid-points to their respective sources \small \mbox{$\pi_{A}, \pi_{B} \in \Pi$} \normalsize with a gradient descent on \small $\mathcal{U}$ \normalsize (\cref{fig2}). The associated integral geodesic length \small $U_{\pi} = \int_{\pi_{A}}^{\pi_{B}}\mathcal{U}d\pi$ \normalsize is computed and the connectivity in \small $\Pi$ \normalsize is updated in the form of an adjacency list. Lastly, the grid-points of the extracted \small $\pi$ \normalsize are further considered as path seeds in the updating scheme, since furcations can occur at any level of the connecting minimal paths.

\textit{Fast Updating Scheme.~}A nested \textit{aFM} is run only in the union of the collided regions \small $(A \cup B)$ \normalsize using a temporary independent layer of \textit{aFM} maps, where \small \mbox{$\tilde{\mathcal{U}}(\pi) = 0$}, \mbox{$\tilde{\mathcal{T}}(\pi) = \text{\textit{Front}}$}, \normalsize and \small \mbox{$\tilde{\mathcal{T}}_{(\overline{A \cup B})} = \text{\textit{Visited}}$}. \normalsize Ideally, the nested \textit{aFM} is run until complete domain exploration, however, to speed up the process, the propagation domain is divided into the solved and \textbf{\textit{unsolved}} sub-regions, and the update is focused on the latter \small $(A \cup B)^{\text{\textbf{\textit{u}}}}$ \normalsize (\cref{fig2}). The boundary geodesic values of \small $(A \cup B)^{\text{\textbf{\textit{u}}}}$ \normalsize equal the geodesic distances \small $\mathcal{U}$ \normalsize at the collision grid-points. Lastly, the \textit{aFM} maps are updated as: \small $\mathcal{U}_{(A \cup B)^{\text{\textbf{\textit{u}}}}} = \min \, \{ \, \mathcal{U}_{(A \cup B)^{\text{\textbf{\textit{u}}}}} , \, \tilde{\mathcal{U}}_{(A \cup B)^{\text{\textbf{\textit{u}}}}} \}$, $\mathcal{V}_{(A \cup B)} = \min \, \{ \, \mathcal{V}_{A} , \, \mathcal{V}_{B} \}$, \normalsize and \small $\mathcal{T}_{(A \cup B)^{\text{\textbf{\textit{u}}}}}= \tilde{\mathcal{T}}_{(A \cup B)^{\text{\textbf{\textit{u}}}}}$. \normalsize
%
\section{Experiments and Results}
\subsubsection{Dataset.}
A 3D hand-crafted tortuous and convoluted phantom \small (HCP) \normalsize is designed to account for complex vessel patterns, i.e. branching, kissing vessels, scale and shape variations induced by pathologies. Also a set of 20 synthetic vascular trees \small (SVT) ($64\times64\times64$ voxels) \normalsize were generated using VascuSynth \cite{vascusynth} considering two levels of additional noise \small (N\textsubscript{1}: $\mathcal{N}(0,5)$\,+\,Shadows: 1\,+\,\mbox{Salt$\&$Pepper: $1\permil$}; ~ N\textsubscript{2}: $\mathcal{N}(0,10)$\,+\,Shadows: 1\,+\,\mbox{Salt$\&$Pepper: $2\permil$}). \normalsize Together with the synthetic data, a cerebral Phase Contrast MRI \small (PC) ($0.86\times0.86\times1.0$ mm), \normalsize a cerebral Time of Flight MRI \small (TOF) ($0.36\times0.36\times0.5$ mm) \normalsize and a carotid \small CTA ($0.46\times0.46\times0.45$ mm) \normalsize were used. Vascular network ground-truths \small (GT) \normalsize are given in the form of connected raster centerlines for all the synthetic images and for both \mbox{\small TOF \normalsize and \small CTA.} \normalsize
\subsubsection{Experiments.}
The scalar vesselness responses of both \small HCP \normalsize and \small PC \normalsize images are determined using the state-of-the-art Frangi filter \small (FFR) \normalsize \cite{frangi1998multiscale}, and Optimally Oriented Flux \small (OOF) \normalsize \cite{law2008three}. Also, the connected vesselness map \small ($\text{\textit{CVM}}$) \normalsize and the associated tensor field \small ($\text{\textit{TF}}$) \normalsize are synthesized for the same dataset using VTrails. The connectedness of the considered scalar maps is qualitatively assessed and the \small $\text{\textit{TF}}$ \normalsize is inspected as proof of concept in section 3.1.\\
VTrails is used to extract the connected geodesic paths for all the synthetic \small SVT \normalsize and for \small TOF \normalsize and \small CTA \normalsize images. In section 3.2, each set of connected geodesic paths is verified to be an acyclic graph, then it is compared against the respective \small GT. \normalsize The robustness to image degradation, the accuracy, precision and recall are evaluated voxel-wise for the identified branches with a tolerance factor \small $\varrho$ \normalsize as in \cite{Annunziata2015}.
\begin{figure}[tb!]
\begin{tabular}{cccccc}
	& \tiny{Image} & \tiny{FFR} & \tiny{OOF} & \tiny{CVM} & \tiny{TF} \\
	\raisebox{-.5\height}{\begin{sideways} \tiny{HCP}\end{sideways}} &
	\raisebox{-.5\height}{\includegraphics[scale=0.072]{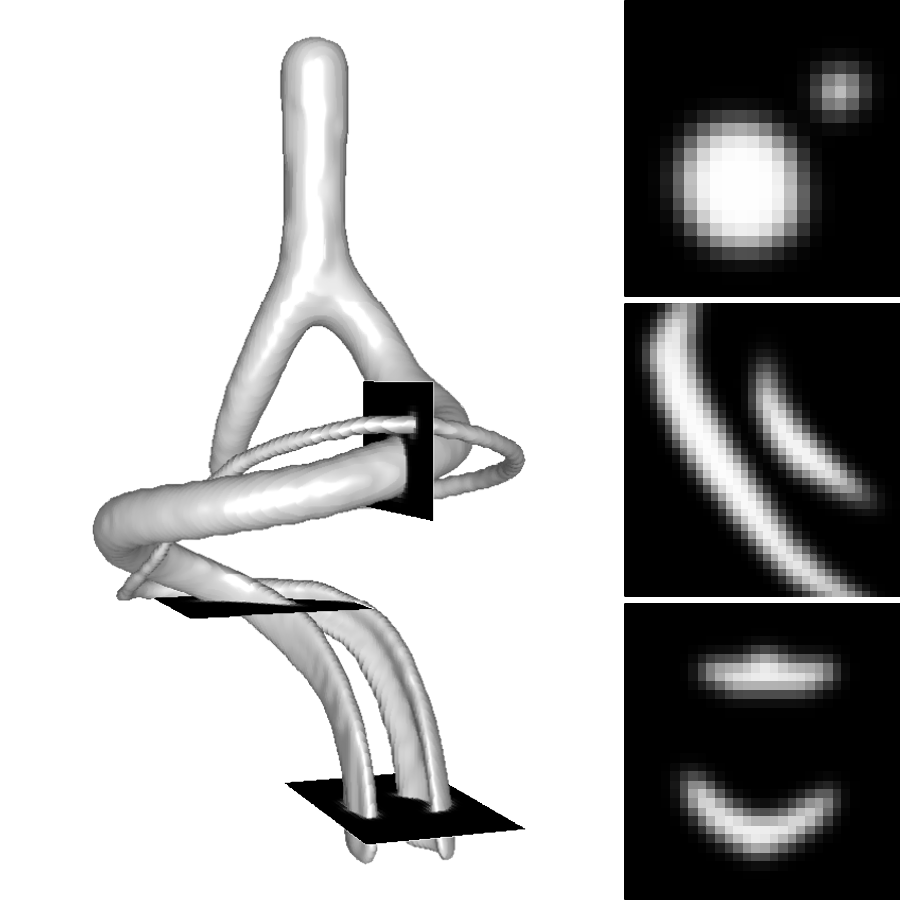}} &
	\raisebox{-.5\height}{\includegraphics[scale=0.072]{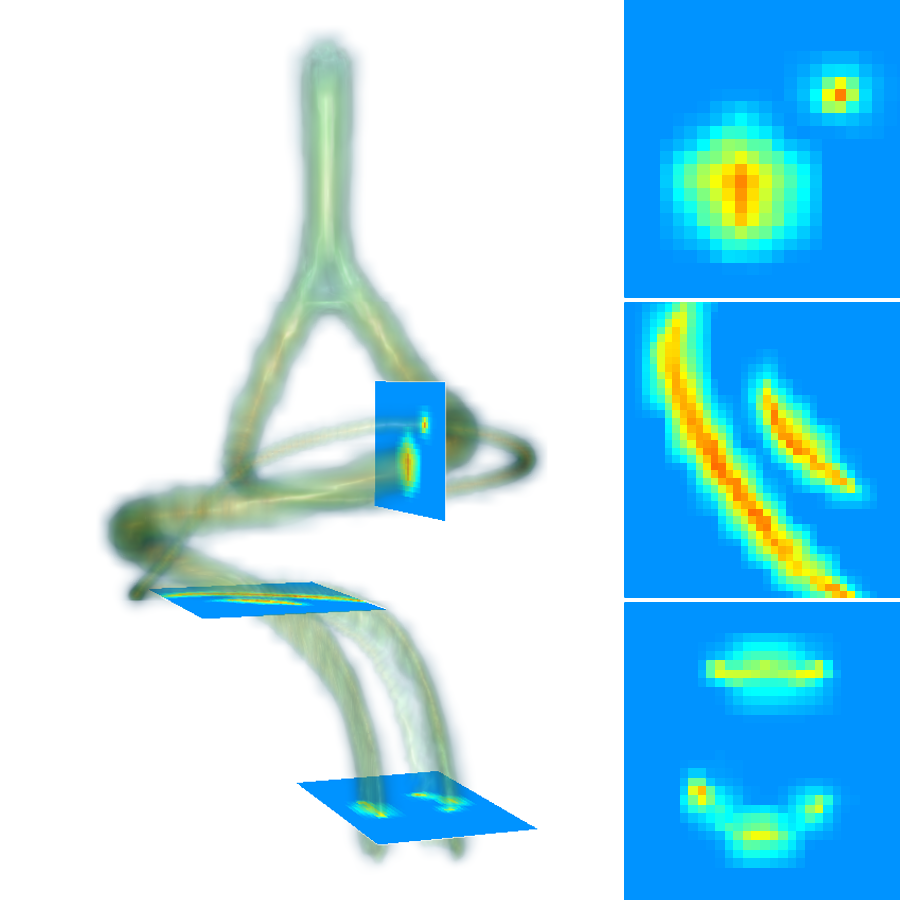}} &
	\raisebox{-.5\height}{\includegraphics[scale=0.072]{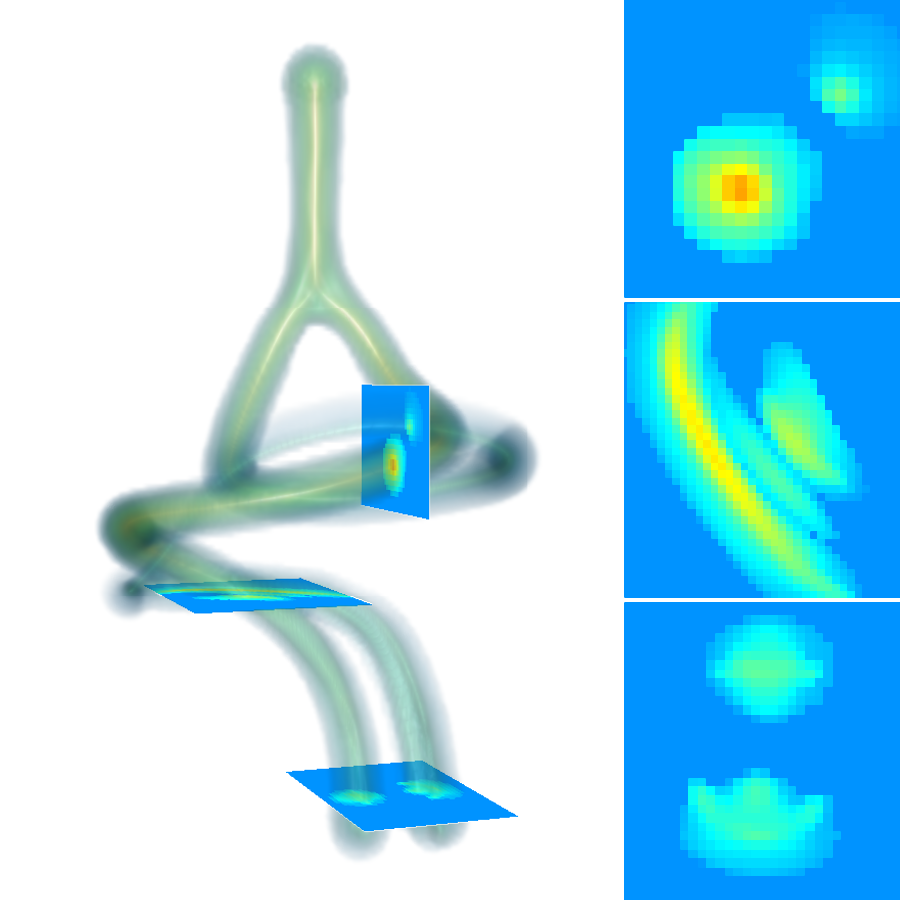}} &
	\raisebox{-.5\height}{\includegraphics[scale=0.072]{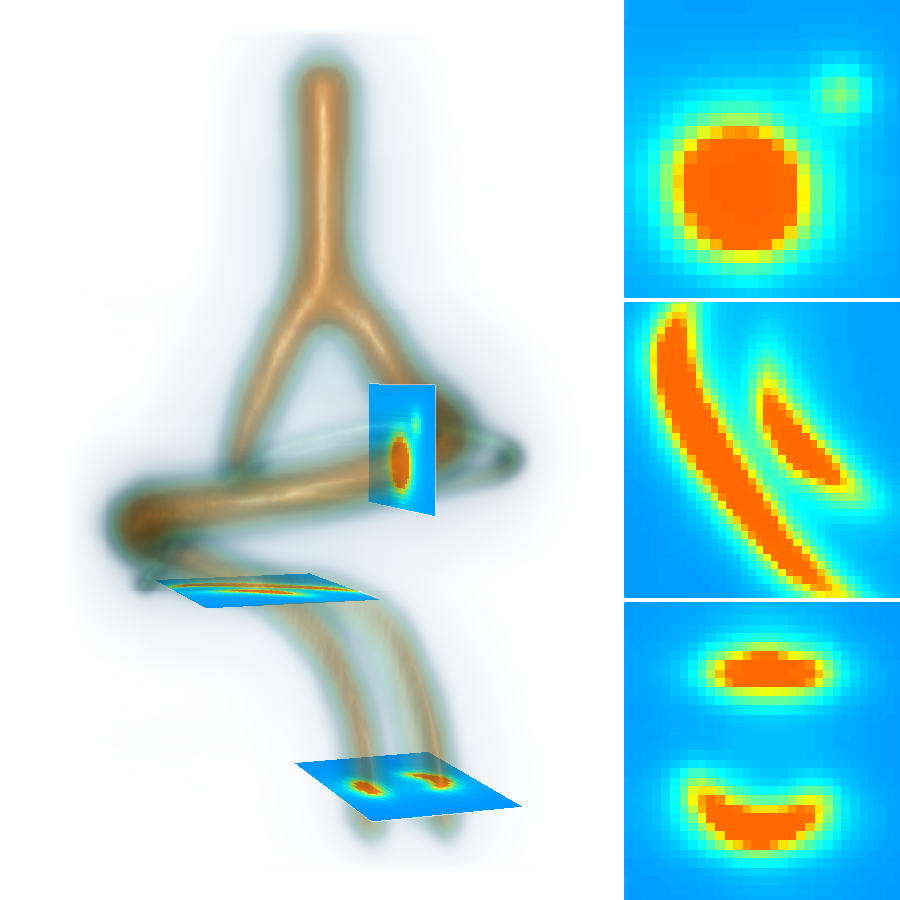}} &
	\raisebox{-.5\height}{\includegraphics[scale=0.072]{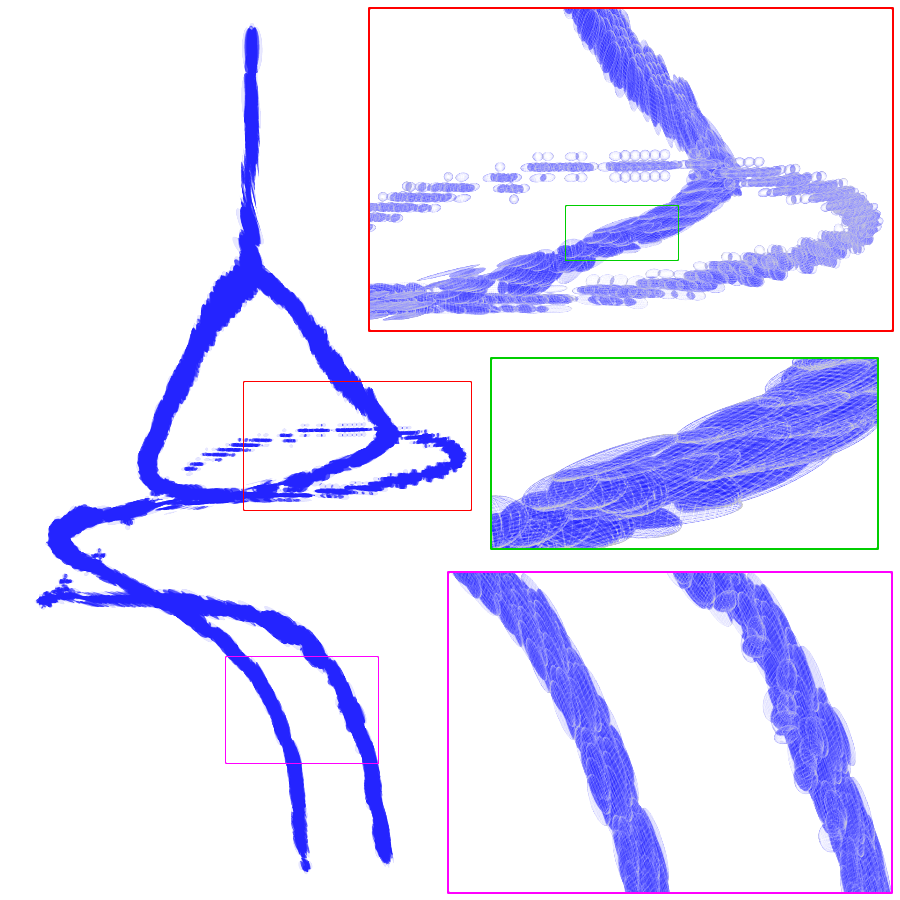}} \\ 
	\raisebox{-.5\height}{\begin{sideways} \tiny{PC}\end{sideways}} &
	\raisebox{-.5\height}{\includegraphics[scale=0.131]{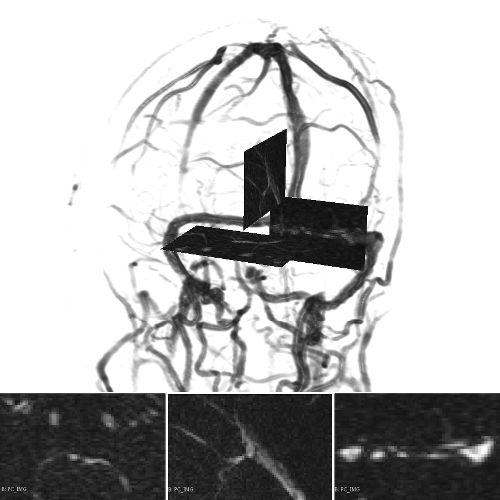}} &
	\raisebox{-.5\height}{\includegraphics[scale=0.131]{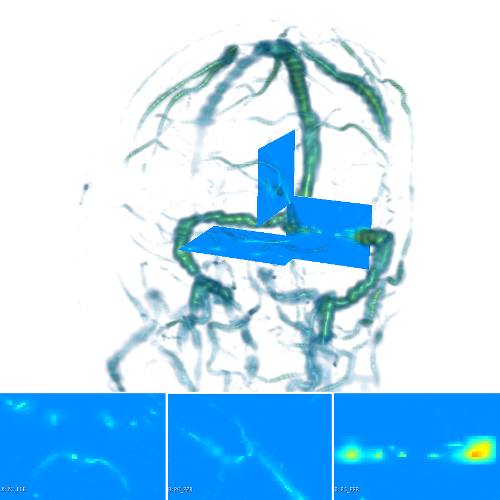}} &
	\raisebox{-.5\height}{\includegraphics[scale=0.131]{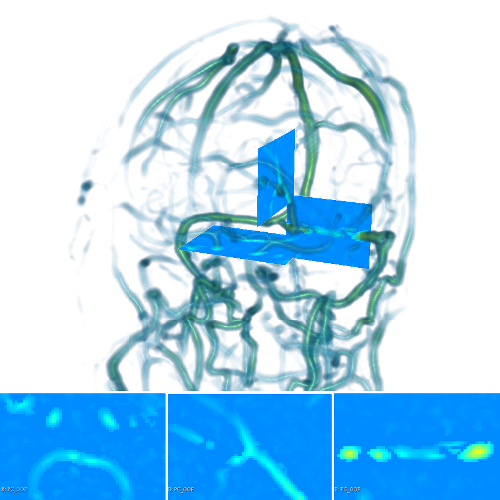}} &
	\raisebox{-.5\height}{\includegraphics[scale=0.131]{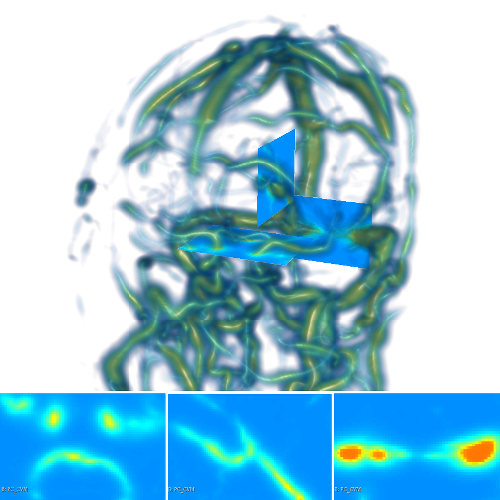}} &
	\raisebox{-.5\height}{\includegraphics[scale=0.072]{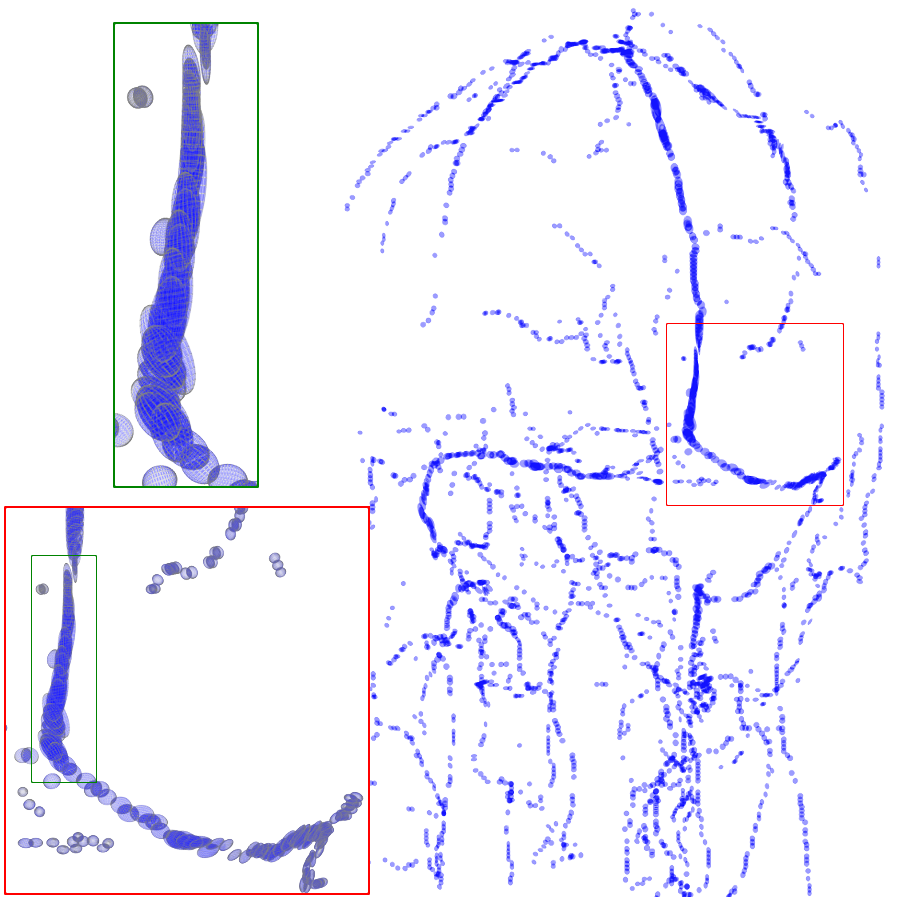}} 
\end{tabular}
\caption{{Vesselness response maps for Frangi, OOF, and proposed scalar CVM with associated tensor field on a digital phantom example and on data of a phase contrast cerebral venogram.}}
\label{fig3}
\end{figure}
\subsection{Connectedness of the Vesselness Map}
Fig. 3 shows the connectedness of vessels recovered from state-of-the-art vascular enhancers and curvilinear ridge detectors \small FFR \normalsize and \small OOF \normalsize together with the proposed \small $\text{\textit{CVM}}$ \normalsize for the synthetic \small HCP \normalsize and the real \small PC \normalsize images. 
On the synthetic phantom, \small FFR \normalsize shows a fragmented and rough vesselness response in correspondence of irregularly shaped sections of the structure. Also, the response at the bifurcation is not smoothly connected with the branches (triangular loop). Conversely, \small OOF \normalsize recovers the phantom connectedness at the branch-point, and the vesselness response is consistent along the tortuous curvilinear section, however ghosting artifacts are observed as the shape of the phantom becomes irregular \small \mbox{(C-like)} \normalsize or differs from a cylindrical tube. Also, close convoluted structures, which change scale rapidly in the \small HCP, \normalsize produce inconsistent responses of \small OOF \normalsize (\cref{fig3}). \small $\text{\textit{CVM}}$ \normalsize shows here a strongly connected vesselness response in correspondence of both regular and irregular tubular sections, with local maxima at structures' mid-line. The connectedness of the structures is emphasized regardless the complexity of the shape, and it resolves spatially the tortuous curvilinear `kissing vessels' without additional ghosting artifacts, despite the smooth profile.\\
Similar results are observed on the \small PC \normalsize dataset: \small FFR \normalsize has a poor connected response in the noisy and low-resolution image. Vessels are overall enhanced, however thin and fragmented structures remain disconnected. Overall, the vesselness response is not uniform within the noisy structures, where maximal values are often off-centred. A more consistent response is obtained from \small OOF, \normalsize where the connectedness of vessels is improved. Maximal response is observed at the mid-line of vessels, however, noise rejection is poor. \small $\text{\textit{CVM}}$ \normalsize strongly enhances here the vessel connectivity. The fragmented vessels of \small PC \normalsize have a continuous and smooth response in \small $\text{\textit{CVM}}$ \normalsize with higher values and a more defined profile. Large vessels shows solid connected regions with local maxima at mid-line as in \small OOF. \normalsize Conversely from \small OOF, $\text{\textit{CVM}}$ \normalsize shows improved noise rejection in the background.\\
The respective tensor fields \small ($\text{\textit{TF}}$) \normalsize synthesized on both \small HCP \normalsize and \small PC \normalsize show consistent features. The \small $\text{\textit{TF}}$'s \normalsize characteristics are in line with the connectedness of \small $\text{\textit{CVM}}$: \normalsize enhanced and connected vessels are associated with high anisotropy, whereas background areas show a predominant isotropic component.
\subsection{Connected Geodesic Paths as Vascular Tree}
\begin{figure}[tb!]
\begin{tabular}{cccc||cccc}
	\multicolumn{4}{c}{\tiny{ \textbf{Synthetic Vascular Trees (SVT)} \cite{vascusynth} } } & \multicolumn{4}{c}{\tiny{\textbf{Clinical Angiographies}}} \\
	& \tiny{Image} & \tiny{GT} & \tiny{VTrails} & & \tiny{Image} & \tiny{GT} & \tiny{VTrails} \\
	\raisebox{-.5\height}{\begin{sideways} \tiny{N\textsubscript{1}}\end{sideways}} &
	\raisebox{-.5\height}{\includegraphics[scale=0.1]{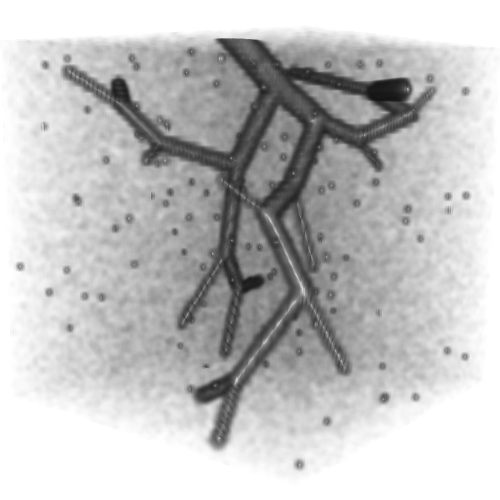} } &
	\raisebox{-.5\height}{\includegraphics[scale=0.2]{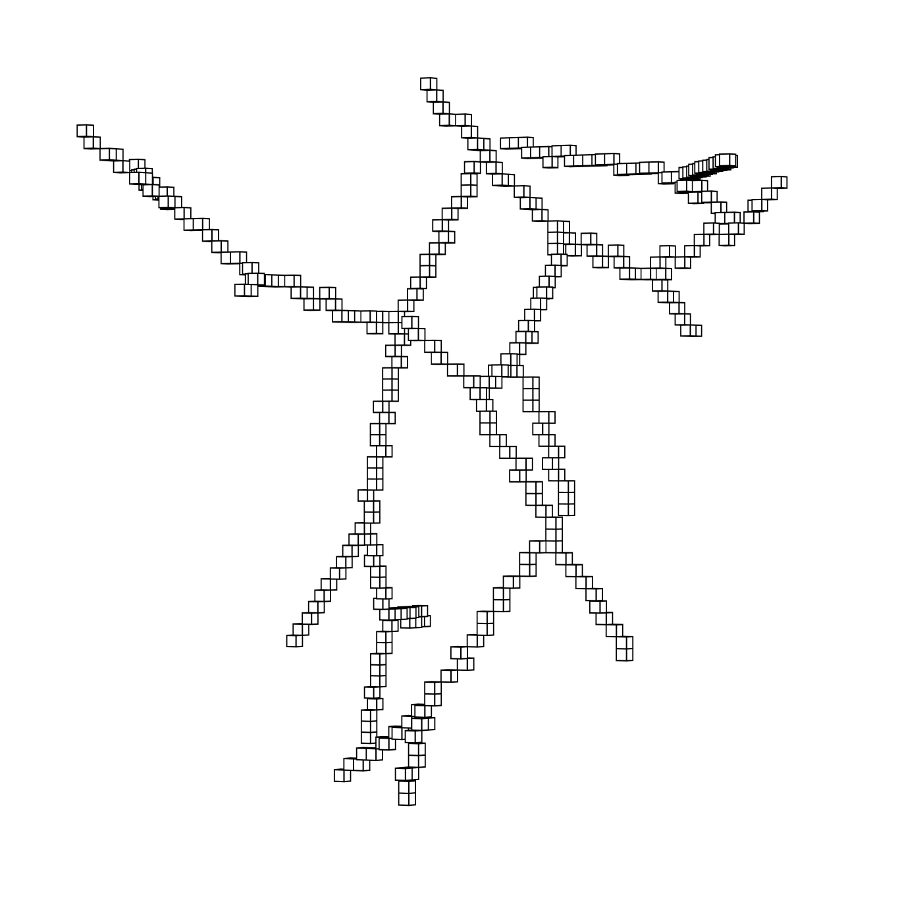} } &
	\raisebox{-.5\height}{\includegraphics[scale=0.2]{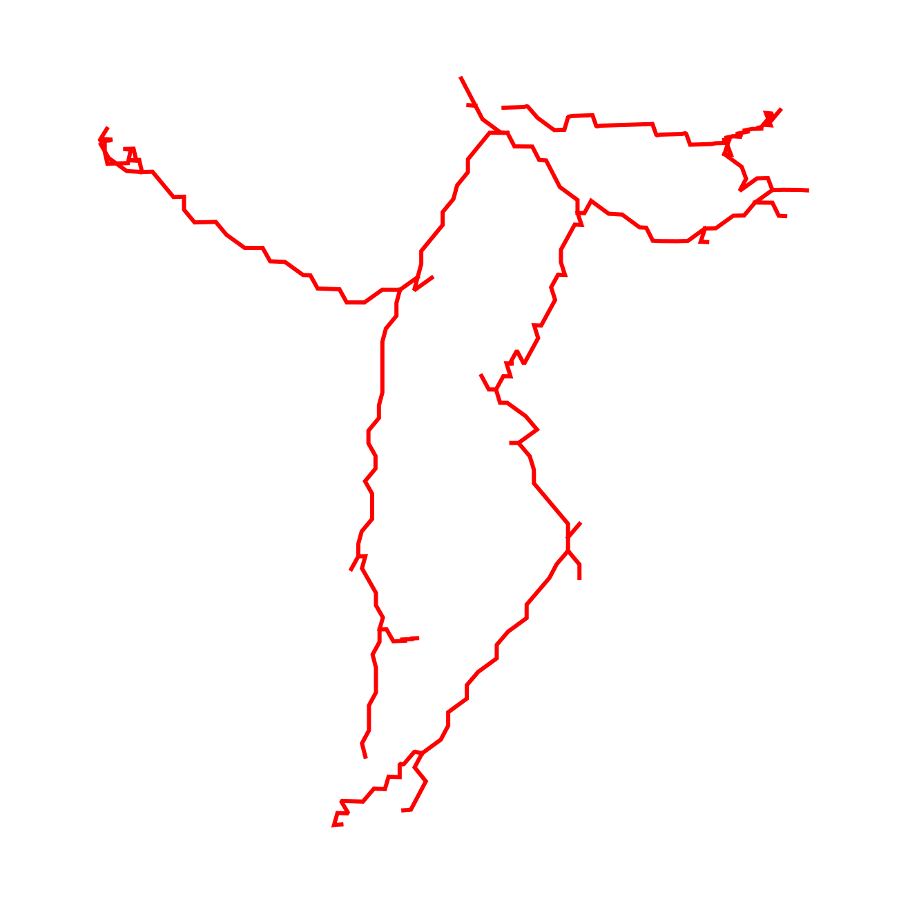} } &
	\raisebox{-.5\height}{\begin{sideways} \tiny{CTA}\end{sideways} } &
	\raisebox{-.5\height}{\includegraphics[scale=0.1]{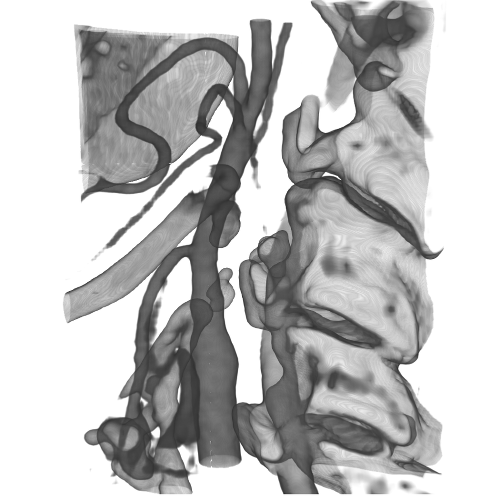} }&
	\raisebox{-.5\height}{\includegraphics[scale=0.2]{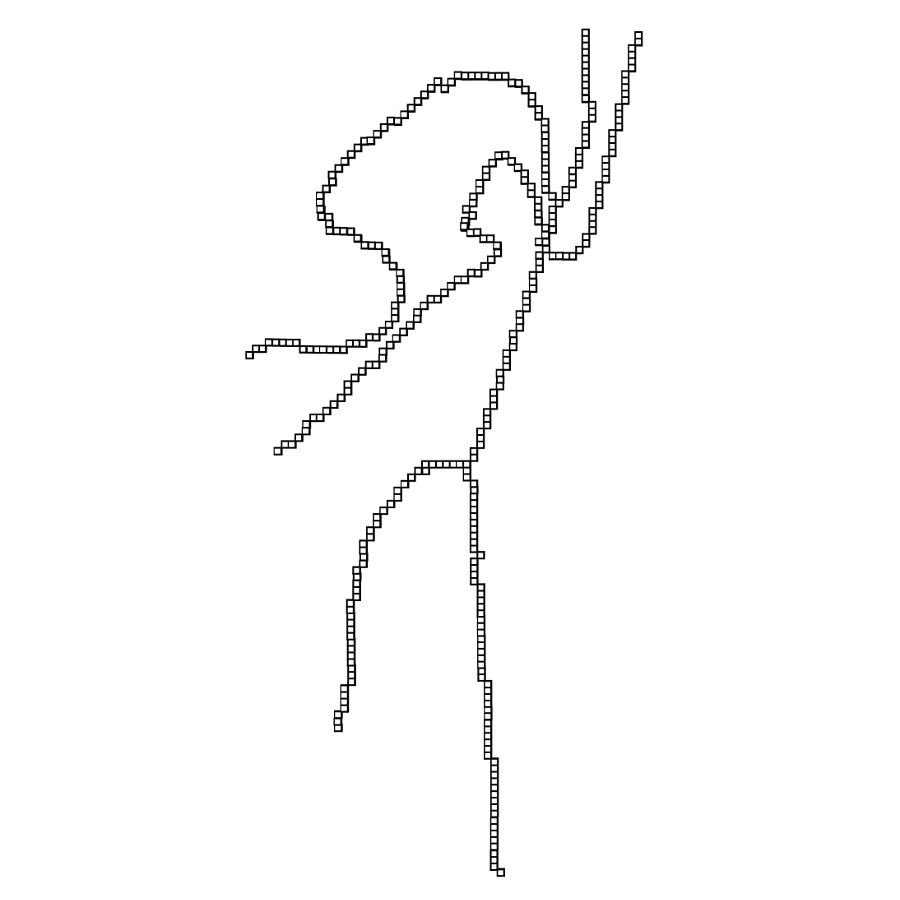} }&
	\raisebox{-.5\height}{\includegraphics[scale=0.2]{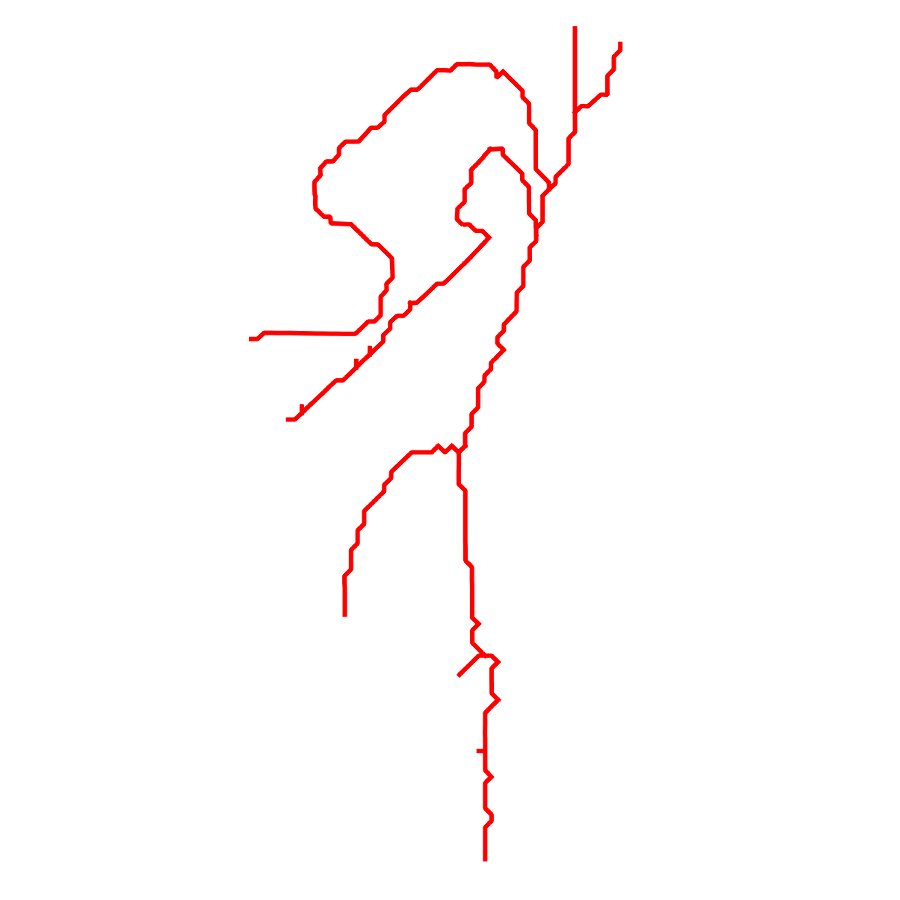} } \\
	\raisebox{-.5\height}{\begin{sideways} \tiny{N\textsubscript{2}}\end{sideways}} &
	\raisebox{-.5\height}{\includegraphics[scale=0.1]{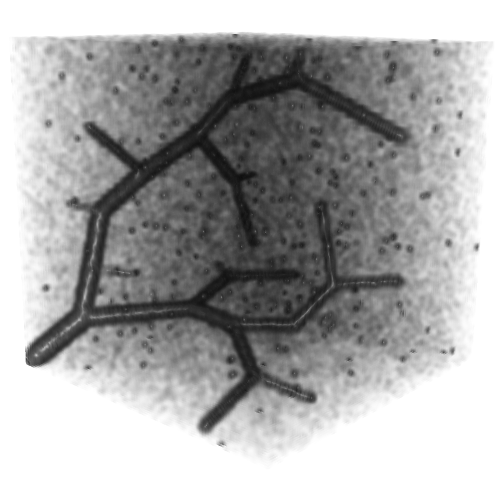} } &
	\raisebox{-.5\height}{\includegraphics[scale=0.2]{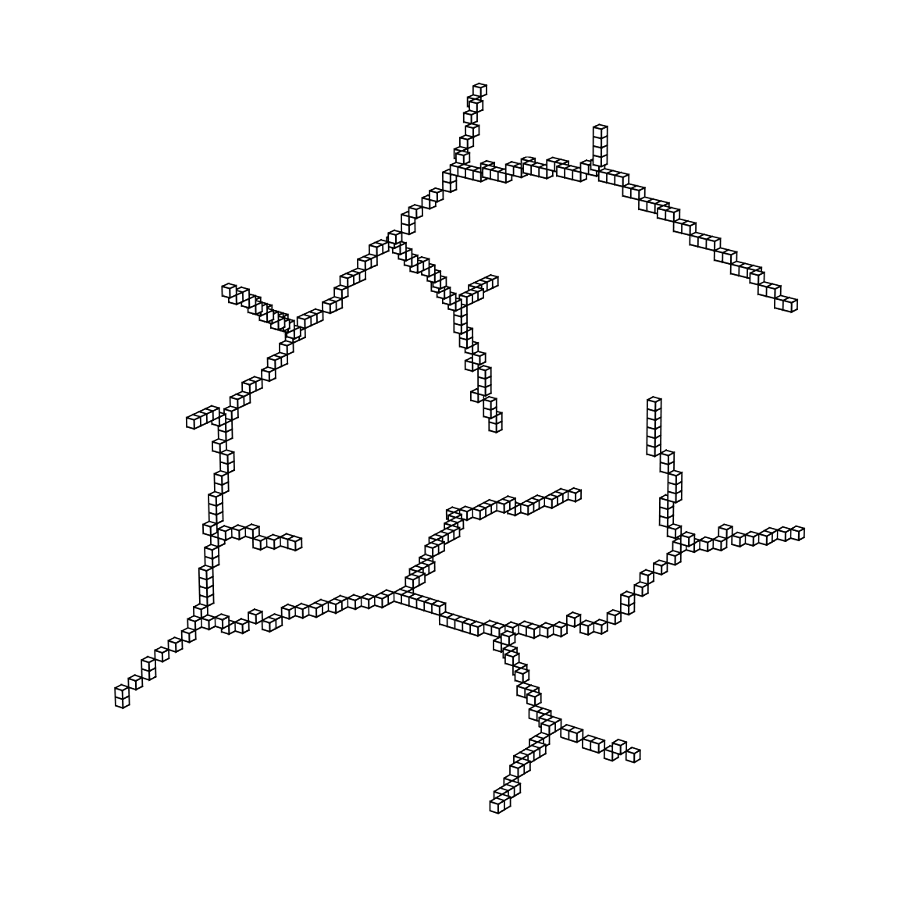} } &
	\raisebox{-.5\height}{\includegraphics[scale=0.2]{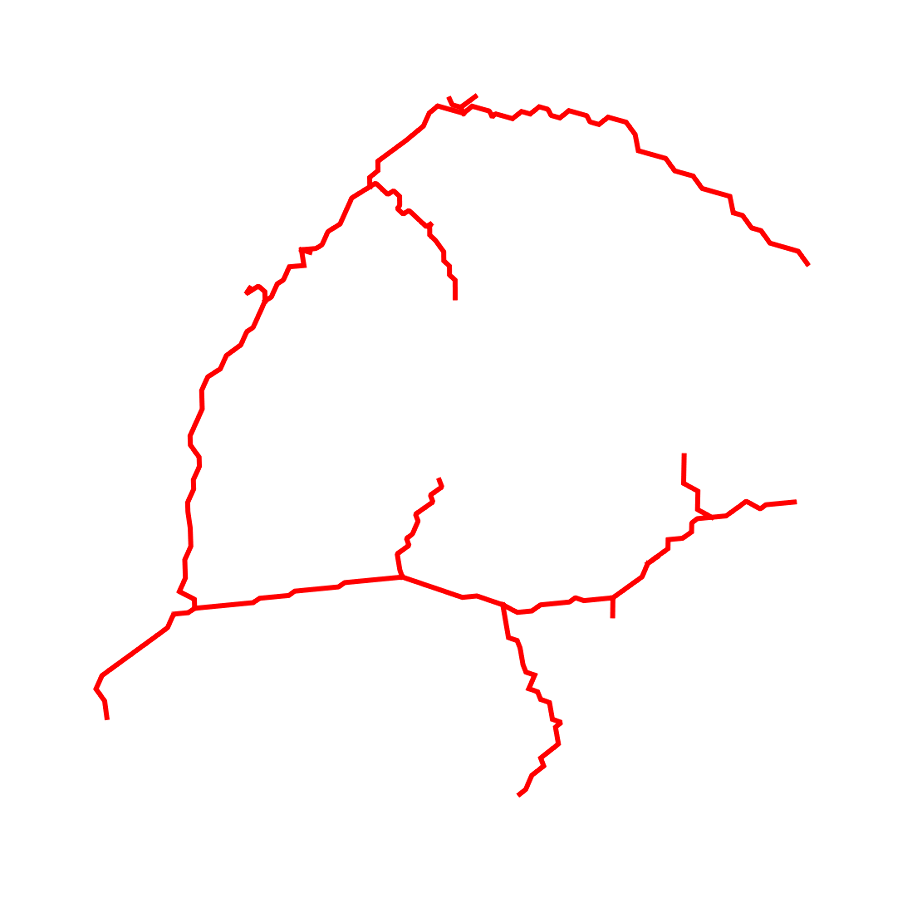} }&
	\raisebox{-.5\height}{\begin{sideways} \tiny{TOF}\end{sideways} } &
	\raisebox{-.5\height}{\includegraphics[scale=0.1]{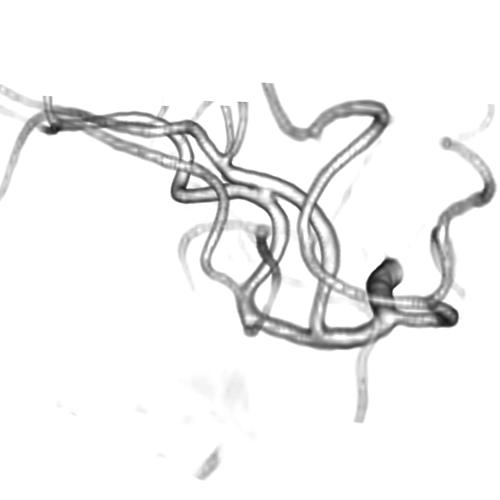} }&
	\raisebox{-.5\height}{\includegraphics[scale=0.2]{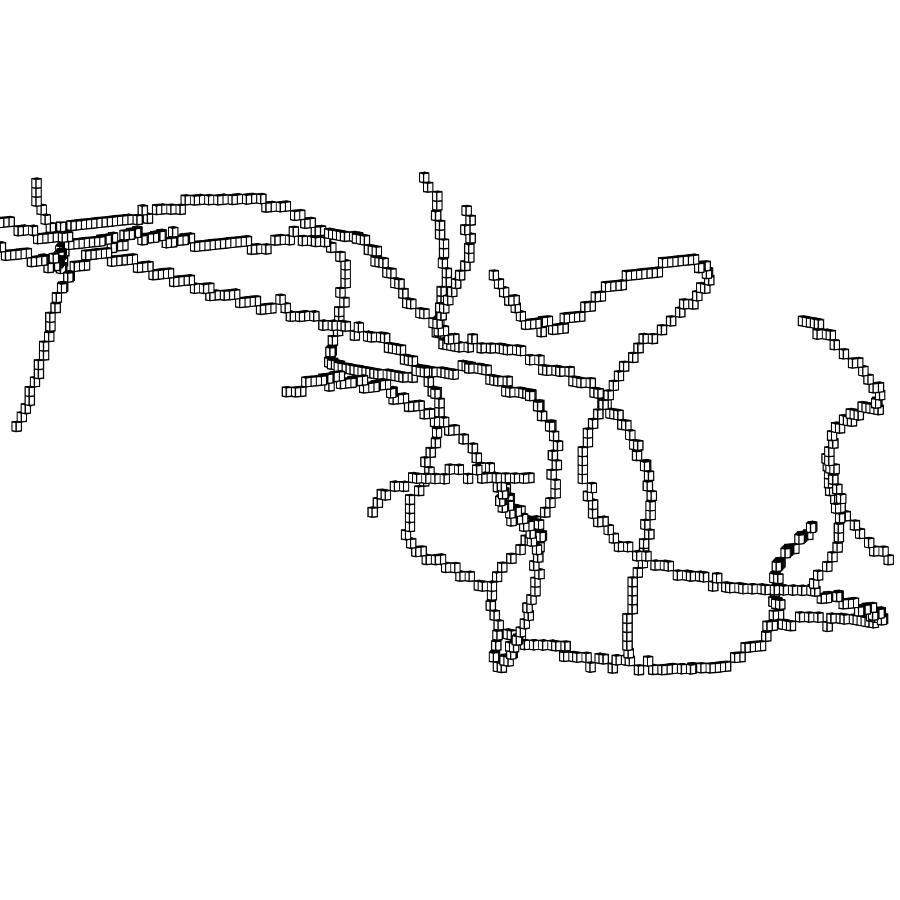} }&
	\raisebox{-.5\height}{\includegraphics[scale=0.2]{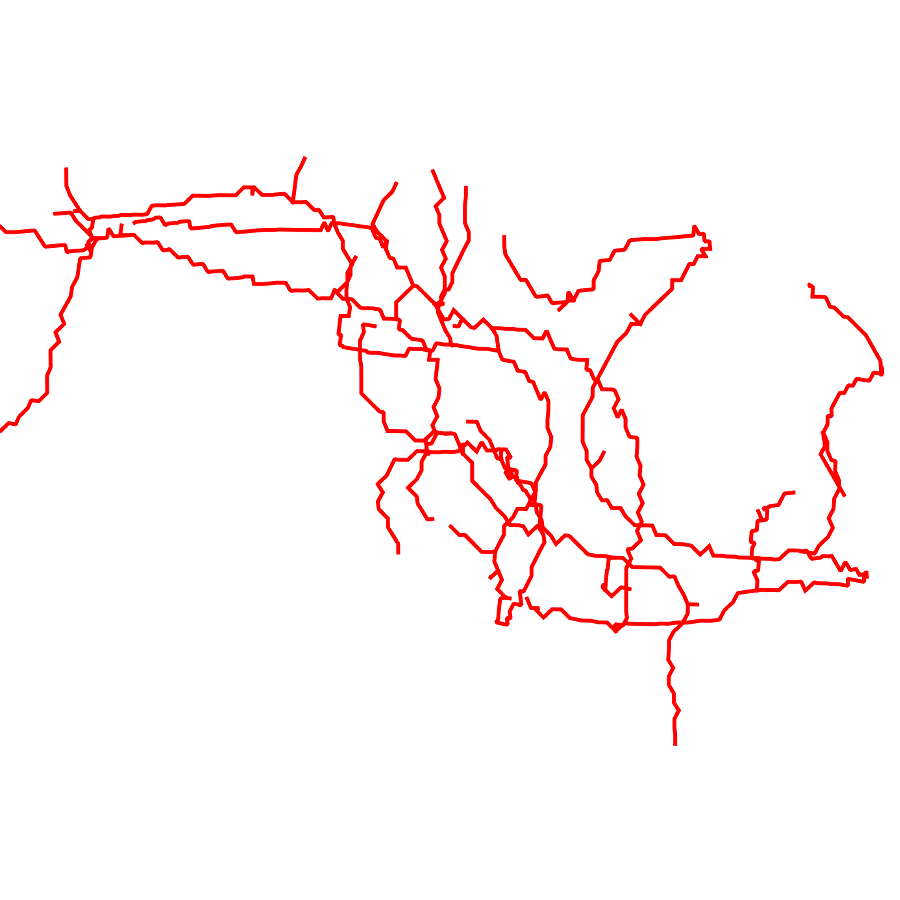} } 

\end{tabular}
\caption{{Comparison of the vascular connected trees against the relative ground-truth for a representative set of synthetic data, and for a carotid CTA and for a middle cerebral artery TOF MRI. Note that main branches are correctly identified and connected.}}
\label{fig4}
\end{figure}

Representative examples of degraded synthetic images from \small SVT \normalsize and the respective \small GT \normalsize are shown in \cref{fig4} together with the connected graphs extracted by VTrails. Analogously, the same set of images are reported for the real images \small TOF \normalsize and \small CTA \normalsize in \cref{fig4}. Qualitatively, the extracted set of connected geodesic paths shows remarkable matching with the provided \small GT \normalsize in all cases.
First, we verify the acyclic nature of the graph. We found no cycles, degenerate graphs and unconnected nodes, meaning that the extracted connected geodesic paths represent a connected geodesic tree.
Precision and recall are then evaluated for the identified branches. Also, error distances are determined as the connected tree's binary distance map evaluated at \small GT. \normalsize Average errors \small ($\overline{\varepsilon}$) \normalsize precision and recall are reported \small (mean$\pm$SD) \normalsize in \cref{tab1}.
Note that no pruning of any spurious branches is performed in the analysis.
%
\section{Discussion and Conclusions}
We presented VTrails, a novel connectivity-oriented method for vascular image analysis. The proposed method has the advantage of introducing the \small SLoGS \normalsize filterbank, which simultaneously synthesizes a connected vesselness map and the associated tensor field in the same mathematically coherent framework. Interestingly, recent works \cite{wang2016tensor, gulsun2016coronary} are exploring Riemannian manifolds of tensors for high-order vascular metrics, however the coherent definition of a tensor field is not trivial for an arbitrary scalar image, as its topology cannot be generally approximated simply by an ellipsoid model \cite{lin2003enhancement}.
The steerability property of \small SLoGS \normalsize stands out as key feature for \textit{i.} reducing the dimensionality of the kernels parameters in 3D, \textit{ii.} determining the filterbank's rotation-invariance and \textit{iii.} optimizing the 3D filtering complexity in the \small OLA. \normalsize Also, the combined rotation- and curvature-invariance of the filtering process results in branch-points that coincide with the locally integrated center of mass of the multiple \small SLoGS \normalsize filter responses.
This explains the strong response in the \small $\text{\textit{CVM}}$ \normalsize at the branch-point in \cref{fig3}.
Regarding the acyclic connectivity paradigm employed in VTrails, we experimentally verified that the resulting set of connected geodesic paths \small $\Pi$ \normalsize forms a tree. The assumption of a vascular tree provides a natural and anatomically valid constraint for 3D vascular images, with few rare exceptions, such as the complete circle of Willis \cite{bergman2007illustrated}. 
It is important to note that the proposed algorithm can include extra anatomical constraints to correct for locations where the vascular topology is not acyclic or where noise it too high. 
Note that despite the optimal formulation of the anisotropic front propagation, a limitation of the greedy acyclic connectivity paradigm is the possibility of miss-connecting branches, potentially disrupting the topology of the vascular network.
Overall, promising results have been reported from this early validation, with a fully-automatic extraction configuration. Missing branches occur in correspondence of small vessels, where the effect of degradation is predominant: tiny terminal vessels completely occluded by the corrupting shadows will not automatically produce seeds, hence cannot be recovered under such configuration. Globally, \small $\overline{\varepsilon}$ \normalsize values are comparable to the evaluation tolerance \small $\varrho$, \normalsize suggesting that the connected geodesic paths extracted by VTrails lie in the close neighbourhood of the vessels' centerlines. Moreover, the reported values are comparable regardless the level of degradation.
Future developments will address the optimization of the \small $\text{\textit{CVM}}$ \normalsize integration strategy in section 2.1 to account for an equalized response over the vascular spatial frequency-bands. Also, the topological analysis of vascular networks on a population of subjects will be investigated in future works to better embed priors in the acyclic connectivity paradigm.
\begin{table}[t]
\begin{tabular}{ccccccc}
	\multicolumn{2}{c}{} & \multicolumn{2}{c}{\tiny{\textbf{Synthetic Vascular Trees} \cite{vascusynth} }} & ~ & \multicolumn{2}{c}{\tiny{\textbf{Clinical Angiographies}}} \\ \cline{3-4} \cline{6-7}
	\multicolumn{2}{c}{} & \scriptsize{N\textsubscript{1}} & \scriptsize{N\textsubscript{2}} & ~ & \scriptsize{TOF} & \scriptsize{CTA} \\
	 $\overline{\varepsilon}$ & \multirow{2}{*}{\begin{sideways}\centering \tiny{[voxels]}\end{sideways}} &
	 \scriptsize{$2.15\pm0.65$} & \scriptsize{$2.09\pm0.37$} & \multirow{2}{*}{\begin{sideways} \centering \tiny{[mm]}\end{sideways}} & \scriptsize{$1.07\pm2.65$~} & \scriptsize{~$1.1\pm1.63$~} \\
	$\varrho$ &  & \multicolumn{2}{c}{\scriptsize{2}} & ~ & \scriptsize{1.42} & \scriptsize{1.57} \\ \hline
	\scriptsize{Precision} &  & \scriptsize{~$88.21\pm2.58\%$~} & \scriptsize{~$87.93\pm2.56\%$~} & ~ & \scriptsize{$77.12\%$} & \scriptsize{$89.67\%$} \\
	\scriptsize{Recall} &  & \scriptsize{$68.31\pm7.44\%$} & \scriptsize{$69.18\pm3.69\%$} & ~ & \scriptsize{$89.49\%$} & \scriptsize{$83.97\%$}
\end{tabular}
\caption{{Connectivity tree error distances, precision and recall (mean$\pm$SD): (left) synthetic vascular tree at degradation levels N\textsubscript{1} and N\textsubscript{2}; (right) TOF and CTA. Note the invariance of all metrics regardless the degradation level.}}
\label{tab1}%
\end{table}%
\section*{Acknowledgements}
The study is co-funded from the EPSRC grant \small (EP/H046410/1), \normalsize the Wellcome Trust and the National Institute for Health Research \small (NIHR) \normalsize University College London Hospitals \small (UCLH) \normalsize Biomedical Research Centre.
%
\bibliographystyle{abbrv}
\bibliography{./IPMI_ID75_CameraReady.bbl}

\begin{thebibliography}{10}

\bibitem{Annunziata2015}
R.~Annunziata, A.~Kheirkhah, P.~Hamrah, and E.~Trucco.
\newblock Scale and curvature invariant ridge detector for tortuous and
  fragmented structures.
\newblock In {\em MICCAI 2015}.

\bibitem{Antiga2008}
L.~Antiga, M.~Piccinelli, L.~Botti, B.~Ene-Iordache, A.~Remuzzi, and D.~A.
  Steinman.
\newblock An image-based modeling framework for patient-specific computational
  hemodynamics.
\newblock {\em Med. Biol. Eng. Comput.}, 2008.

\bibitem{arsigny2006log}
V.~Arsigny, P.~Fillard, X.~Pennec, and N.~Ayache.
\newblock Log-euclidean metrics for fast and simple calculus on diffusion
  tensors.
\newblock {\em Magn. Reson. Med.}, 2006.

\bibitem{Benmansour2011Aniso}
F.~Benmansour and L.~D. Cohen.
\newblock Tubular structure segmentation based on minimal path method and
  anisotropic enhancement.
\newblock {\em Int. J. Comput. Vision}, 2011.

\bibitem{bergman2007illustrated}
R.~A. Bergman, A.~K. Afifi, and R.~Miyauchi.
\newblock Illustrated encyclopedia of human anatomic variation: Circle of
  Willis.
\newblock {\em www.anatomyatlases.org/AnatomicVariants/}.

\bibitem{Bullitt1999}
E.~Bullitt, S.~Aylward, A.~Liu, J.~Stone, S.~K. Mukherji, C.~Coffey, G.~Gerig,
  and S.~M. Pizer.
\newblock 3D graph description of the intracerebral vasculature from segmented
  mra and tests of accuracy by comparison with x-ray angiograms.
\newblock In {\em IPMI 1999}.

\bibitem{cardoso2015scale}
M.~J. Cardoso, M.~Modat, T.~Vercauteren, and S.~Ourselin.
\newblock Scale factor point spread function matching: beyond aliasing in image
  resampling.
\newblock In {\em MICCAI 2015}.

\bibitem{frangi1998multiscale}
A.~F. Frangi, W.~J. Niessen, K.~L. Vincken, and M.~A. Viergever.
\newblock Multiscale vessel enhancement filtering.
\newblock In {\em MICCAI 1998}.

\bibitem{gulsun2016coronary}
M.~A. G{\"u}ls{\"u}n, G.~Funka-Lea, P.~Sharma, S.~Rapaka, and Y.~Zheng.
\newblock Coronary centerline extraction via optimal flow paths and cnn path
  pruning.
\newblock In {\em MICCAI 2016}.

\bibitem{vascusynth}
G.~Hamarneh and P.~Jassi.
\newblock Vascusynth: simulating vascular trees for generating volumetric image
  data with ground-truth segmentation and tree analysis.
\newblock {\em Comput. Med. Imag. Graph.}, 2010.

\bibitem{kwitt2013studying}
R.~Kwitt, D.~Pace, M.~Niethammer, and S.~Aylward.
\newblock Studying cerebral vasculature using structure proximity and graph
  kernels.
\newblock In {\em MICCAI 2013}.

\bibitem{law2008three}
M.~W. Law and A.~C. Chung.
\newblock Three dimensional curvilinear structure detection using optimally
  oriented flux.
\newblock In {\em ECCV 2008}.

\bibitem{Lesage2009}
D.~Lesage, E.~D. Angelini, I.~Bloch, and G.~Funka-Lea.
\newblock A review of 3D vessel lumen segmentation techniques: Models, features
  and extraction schemes.
\newblock {\em Med. Image Anal.}, 2009.

\bibitem{lin2003enhancement}
Q.~Lin.
\newblock Enhancement, extraction, and visualization of 3D volume data.
\newblock 2003.

\bibitem{oppenheim2010discrete}
A.~V. Oppenheim and R.~W. Schafer.
\newblock {\em Discrete-time signal processing}.
\newblock Pearson Higher Education, 2010.

\bibitem{schaap2007bayesian}
M.~Schaap, R.~Manniesing, I.~Smal, T.~Van~Walsum, A.~Van Der~Lugt, and
  W.~Niessen.
\newblock Bayesian tracking of tubular structures and its application to
  carotid arteries in cta.
\newblock In {\em MICCAI 2007}.

\bibitem{wang2016tensor}
C.~Wang, M.~Oda, Y.~Hayashi, Y.~Yoshino, T.~Yamamoto, A.~F. Frangi, and
  K.~Mori.
\newblock Tensor-based graph-cut in riemannian metric space and its application
  to renal artery segmentation.
\newblock In {\em MICCAI 2016}.

\bibitem{Zuluaga2015}
M.~A. Zuluaga, R.~Rodionov, M.~Nowell, S.~Achhala, G.~Zombori, A.~F. Mendelson,
  M.~J. Cardoso, A.~Miserocchi, A.~W. McEvoy, J.~S. Duncan, and S.~Ourselin.
\newblock Stability, structure and scale: improvements in multi-modal vessel
  extraction for seeg trajectory planning.
\newblock {\em Int. J. Comput. Assist. Radiol. Surg.}, 2015.

\end{thebibliography}

\end{document}